\documentclass[10pt,twocolumn,letterpaper]{article}

\usepackage[pagenumbers]{cvpr}

\usepackage{tabularx}   
\usepackage{placeins}   
\usepackage{float}      
\usepackage{multirow}
\usepackage{cuted}

\usepackage[table]{xcolor}
\usepackage{booktabs}
\usepackage{amssymb}

\definecolor{vigerTableGroup}{RGB}{228,239,245}
\definecolor{vigerTableOurs}{RGB}{244,248,251}
\definecolor{vigerOnline}{RGB}{51,123,81}
\definecolor{vigerOffline}{RGB}{164,71,65}

\newcommand{\vigerYes}{%
  \textcolor{vigerOnline}{$\checkmark$}}
\newcommand{\vigerNo}{%
  \textcolor{vigerOffline}{$\times$}}

\newcommand{\vigerMainTableWidth}{0.96\textwidth}

\newcolumntype{Z}{>{\centering\arraybackslash}X}

\newcommand{\vigerMainTableStyle}{%
  \fontsize{7.0pt}{7.7pt}\selectfont
  \setlength{\tabcolsep}{2.0pt}%
  \renewcommand{\arraystretch}{0.94}%
}


\newcommand{\qualcasefigure}[3]{%
    \begin{figure*}[!t]
        \centering
        \setlength{\abovecaptionskip}{3pt}
        \setlength{\belowcaptionskip}{0pt}
        \includegraphics[width=0.90\textwidth]{#1}
        \caption{#2}
        \label{#3}
    \end{figure*}
}

\definecolor{cvprblue}{rgb}{0.21,0.49,0.74}
\usepackage[pagebackref,breaklinks,colorlinks,allcolors=cvprblue]{hyperref}

\def\confName{CVPR}
\def\confYear{2027}

\title{GeoRecall: Streaming Spatial Reasoning through Cross-Level Geometry and Historical Evidence Re-reading}

\author{
Tianyu Geng$^{1}$,
Yuhan Liu$^{2}$,
Jianning Chi$^{1}$,
Hansen Zhang$^{1}$,
Yifan Zhao$^{1}$,
Pei Fu$^{2}$\\
$^{1}$Northeastern University
\quad
$^{2}$Xiaomi, SeerRay Team\\
{\tt\small gengtianyu@mails.neu.edu.cn}
}

\begin{document}
\maketitle

\begin{strip}
    \centering
    \includegraphics[width=0.96\textwidth]{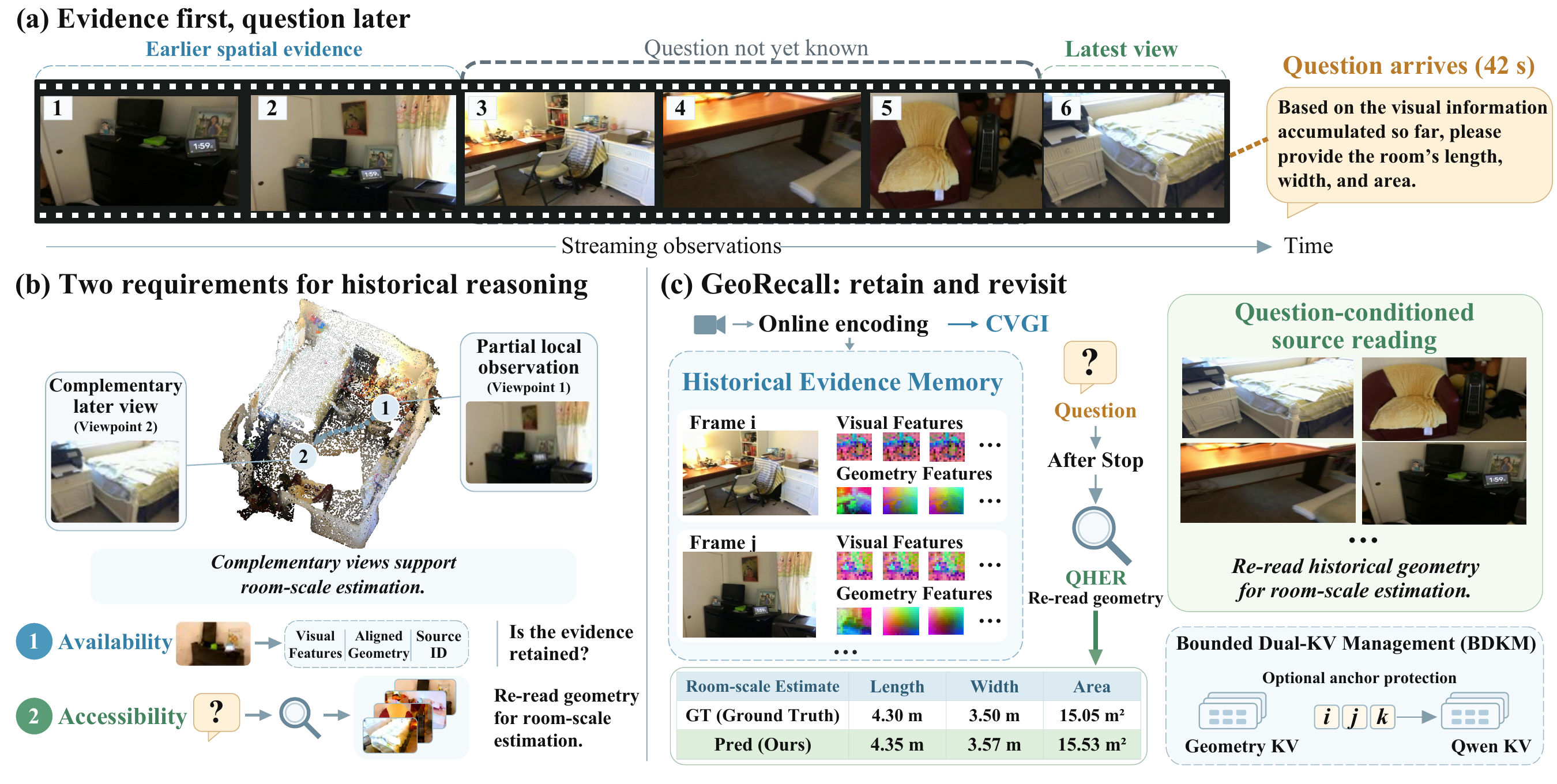}

    \captionsetup{hypcap=false}
    \captionof{figure}{
        Motivation of GeoRecall.
        (a) Questions may arrive after relevant observations have entered the stream.
        (b) Historical reasoning requires evidence retention and question-conditioned access.
        (c) GeoRecall retains source-indexed visual--geometric evidence for historical re-reading, with optional bounded dual-KV management.
    }
    \label{fig:motivation}
    \captionsetup{hypcap=true}
\end{strip}
\begin{abstract}
Streaming spatial question answering requires vision-language models (VLMs) to reason over causally observed RGB streams, where relevant spatial evidence may appear before a query arrives. Despite advances in geometry-enhanced VLMs, integrating multilevel geometric information during streaming observation and accessing corresponding historical geometry at answer time remain challenging. To address these challenges, we propose GeoRecall, a streaming spatial reasoning framework that couples observation-time multilevel visual--geometric interaction with answer-time historical evidence re-reading. It comprises two complementary mechanisms: Cross-Level Visual--Geometric Interaction (CVGI) and Question-Conditioned Historical Evidence Re-reading (QHER). CVGI injects multilevel causal geometric features into evolving visual representations at multiple language-model layers. QHER decouples question-agnostic evidence preservation during streaming observation from question-conditioned retrieval at answer time. Specifically, when answer generation is triggered, the current question guides the construction of source-bound Evidence Tokens from retained observations. Geometry Readers then use these tokens to progressively access the corresponding multilevel geometric features retained in memory, refining the representation of historical spatial evidence without re-encoding past frames. As an auxiliary, training-free inference extension, Bounded Dual-KV Management (BDKM) independently manages the Geometry KV and Qwen KV caches under separate capacity budgets during long-stream inference. Experiments across four benchmarks show that GeoRecall achieves state-of-the-art overall performance on Stream3D-Bench and competitive results on three additional spatial benchmarks. These results highlight the effectiveness of integrating multilevel geometry during streaming observation and question-conditioned re-reading of historical geometric evidence for causal spatial reasoning in streaming environments.
\end{abstract}
\section{Introduction}
\label{sec:intro}

Spatial question answering from live RGB streams requires
vision--language models (VLMs) to reason over observations
as they arrive, rather than after a complete video is
available. As illustrated in Fig.~\ref{fig:motivation}(a),
relevant spatial evidence may precede a question, while
later observations may no longer contain the corresponding
view. Models must therefore construct spatial representations
causally and preserve historical evidence for later reasoning.

Recent 3D models, including DUSt3R~\cite{wang2024dust3r},
VGGT~\cite{wang2025vggt}, CUT3R~\cite{wang2025cut3r}, and
StreamVGGT~\cite{zhuo2026streamvggt}, make geometric
representations increasingly accessible from RGB inputs.
Geometry-enhanced VLMs, including VLM-3R~\cite{fan2026vlm3r},
SpaceMind~\cite{zhao2026spacemind},
SpatialStack~\cite{zhang2026spatialstack}, and
GUIDE~\cite{wang2026guide}, explore visual--geometric
integration; GeoWeaver~\cite{miao2026geoweaver},
GeoThinker~\cite{li2026geothinker}, and
GeoReader~\cite{li2026georeader} study geometry-grounded
reasoning. Stream3D-VLM~\cite{yu2026stream3dvlm} and
SpaTime~\cite{yin2026spatime} extend geometric reasoning to
sequential observations. However, integrating multilevel
geometry into evolving visual representations across multiple
language-model layers under causal streaming constraints
remains challenging.

Beyond online geometric integration, streaming models must
also preserve historical evidence and retrieve its
corresponding geometry for later questions.
VideoLLM-online~\cite{chen2024videollmonline} studies online
video inference; ReKV~\cite{di2025rekv},
StreamKV~\cite{chen2026streamkv}, and
MuKV~\cite{xiao2026mukv} explore historical representation
retrieval. FluxMem~\cite{xie2026fluxmem},
VLM$^2$~\cite{liu2026vlm2}, and
Q-GeoMem~\cite{gao2026qgeomem} investigate visual or
geometric memory. As shown in Fig.~\ref{fig:motivation}(b),
historical spatial reasoning requires both
\emph{evidence availability} and
\emph{question-dependent accessibility}: evidence must
be retained before a question is known and later accessed
according to that question. This calls for separating
question-agnostic evidence preservation from answer-time
geometric retrieval while maintaining source correspondence
between historical visual evidence and retained geometry.
Long streams also increase memory use in geometric and
language-model KV caches. H2O~\cite{zhang2023h2o},
SnapKV~\cite{li2024snapkv}, and
IncVGGT~\cite{fang2026incvggt} offer cache-management
strategies, but the distinct KV structures and retention
needs of geometry encoders and VLMs motivate separate
policies.

To address these challenges, we propose GeoRecall,
a streaming spatial reasoning framework that couples
observation-time multilevel geometric interaction with
answer-time historical evidence re-reading. As illustrated
in Fig.~\ref{fig:motivation}(c) and Fig.~\ref{fig:framework},
it comprises two complementary mechanisms.
Cross-Level Visual--Geometric Interaction (CVGI) aligns
multilevel causal geometric features and injects them into
evolving visual representations at multiple language-model
layers during streaming observation. Question-Conditioned
Historical Evidence Re-reading (QHER) separates
question-agnostic evidence preservation from answer-time
geometric retrieval. The current question guides the
construction of source-bound Evidence Tokens from retained
observations, which direct Geometry Readers to progressively
access corresponding multilevel geometric features retained
in memory, thereby refining the representation of historical
spatial evidence without re-encoding past frames.
We also introduce Bounded Dual-KV Management (BDKM) as a
separately evaluated, training-free inference extension.
It applies independent budgets to the geometry and Qwen KV
caches, pruning geometry tokens by attention importance and
evicting Qwen frame groups, while protecting jointly selected
visual--geometric anchors only in the Qwen cache.

Our contributions are summarized as follows:
\begin{itemize}
    \item We propose GeoRecall, coupling observation-time
    multilevel geometric interaction with answer-time
    historical evidence re-reading, and introduce BDKM as a
    separately evaluated, training-free extension for
    bounded management of the geometry and Qwen KV caches.

    \item We develop CVGI and QHER as complementary
    mechanisms for multilevel visual--geometric fusion and
    question-conditioned historical geometry access.
    QHER combines question-agnostic evidence preservation,
    source-bound Evidence Tokens, and progressive retrieval
    through multilevel Geometry Readers.

    \item We evaluate GeoRecall on four benchmarks,
    achieving a state-of-the-art overall QA score of 71.4
    on Stream3D-Bench and competitive results on VSI-Bench,
    OST-Bench, and MindCube-Tiny.
\end{itemize}
\section{Related Work}
\label{sec:related_work}

\noindent\textbf{Geometry-Augmented Spatial Reasoning.}
Geometry-enhanced VLMs incorporate geometric representations
to support spatial understanding.
SpatialStack~\cite{zhang2026spatialstack} and
GUIDE~\cite{wang2026guide} progressively inject multilevel
geometry into language-model layers, while
GeoReader~\cite{li2026georeader} learns to read latent
geometric representations.
For streaming applications, geometric integration must
respect causal visibility while updating visual
representations as new observations arrive.

\noindent\textbf{Streaming Spatial Question Answering.}
Streaming video models process evolving observations
and determine when to respond.
VideoLLM-online~\cite{chen2024videollmonline} studies
online video inference.
Stream3D-VLM~\cite{yu2026stream3dvlm} combines incremental
geometry, visual--spatial fusion, and response decisions,
while SpaTime~\cite{yin2026spatime} investigates causal
geometry and response-time supervision.
When relevant observations precede a question, spatial
reasoning additionally requires access to evidence
that is no longer visible in the current view.

\noindent\textbf{Question-Conditioned Historical Evidence Retrieval.}
Historical memory supports reasoning beyond the current
observation. ReKV~\cite{di2025rekv} retrieves historical
video representations, while VLM$^2$~\cite{liu2026vlm2}
investigates visual memory for spatial reasoning.
Q-GeoMem~\cite{gao2026qgeomem} combines question relevance
and bank-relative novelty to guide evidence replacement
and reading in a semantic--geometric memory.
A central consideration is how evidence can be preserved
before future questions are known and later accessed
according to their spatial requirements, while maintaining
correspondence with its source observations.

\noindent\textbf{Memory-Efficient Streaming Video and Geometry.}
Long-stream inference requires controlling the growth
of internal states.
StreamingLLM~\cite{xiao2024streamingllm} and
H$_2$O~\cite{zhang2023h2o} study KV retention,
while IncVGGT~\cite{fang2026incvggt} investigates
memory-efficient streaming geometry.
Combining geometric and language processing introduces
caches with different structures and retention requirements,
making resource control and preservation of useful
historical information a joint challenge.

\section{Method}
\label{sec:method}

\subsection{Overview}
\label{sec:method_overview}

We consider streaming spatial question answering over
sequential RGB observations
$\mathcal I=\{I_t\}_{t=1}^{T}$. A question $q_\tau$
arrives at observation time $\tau$, and answer generation
is triggered at $t_s\geq\tau$ according to the model's
response decision. At $t_s$, only the available causal
observation prefix is accessible, so relevant spatial
evidence may originate from observations both before
and after question arrival.

\begin{figure*}[t]
    \centering
    \IfFileExists{figures/framework.pdf}{%
        \includegraphics[width=0.97\textwidth]{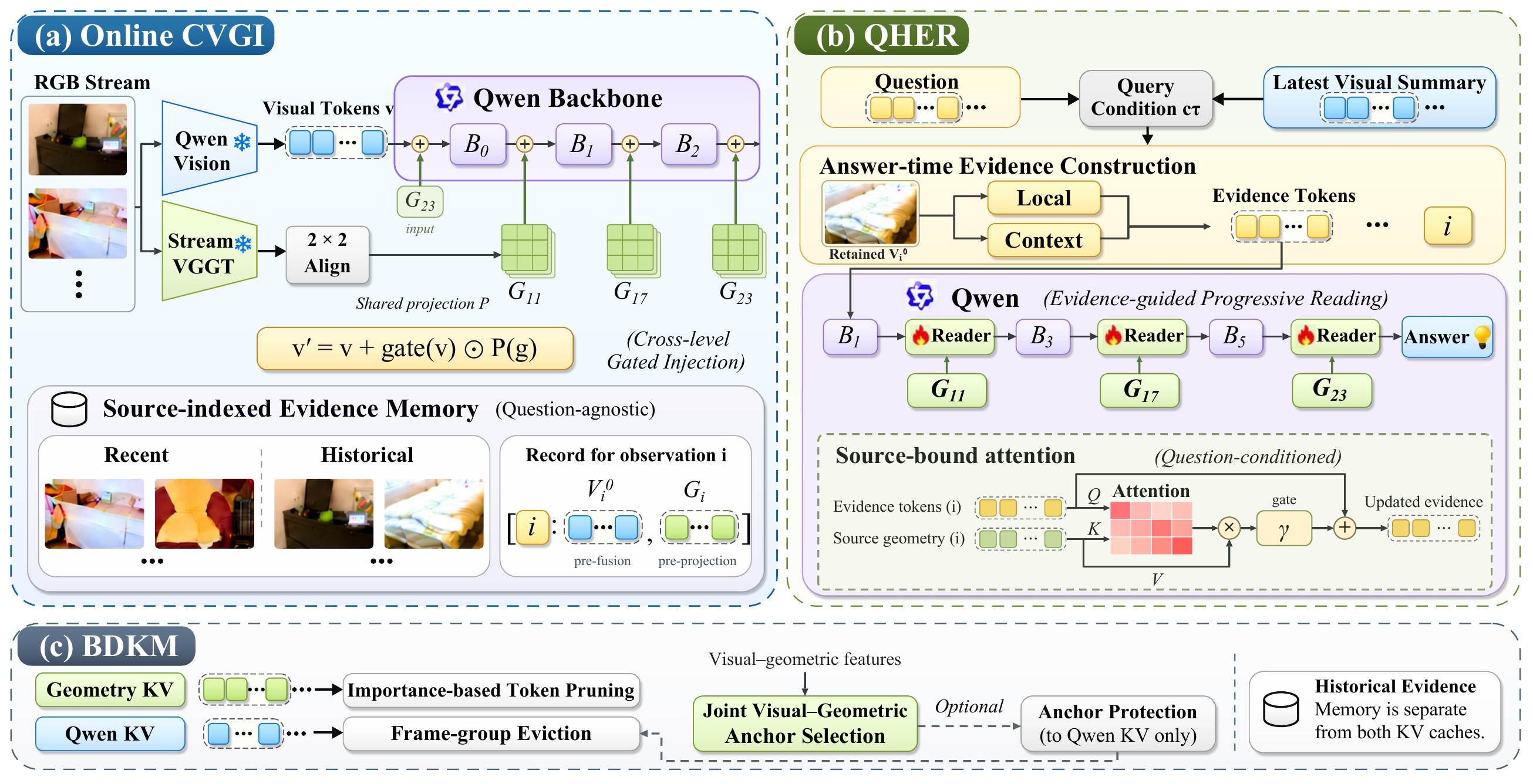}%
    }{%
        \fbox{\parbox[c][0.85in][c]{0.95\textwidth}{\centering
        GeoRecall framework: CVGI, Historical Evidence Memory,
        QHER, and ACDM.}}%
    }
    \caption{Overall architecture of GeoRecall.
(a) CVGI integrates multilevel causal geometry into Qwen
representations while retaining source-indexed evidence.
(b) QHER constructs question-conditioned Evidence Tokens
for source-bound geometry retrieval.
(c) BDKM selects joint visual--geometric frame anchors
and applies cache-specific retention across Geometry
and Qwen KV caches.}
    \label{fig:framework}
\end{figure*}

We propose GeoRecall (Fig.~\ref{fig:framework}), whose
core consists of two complementary mechanisms:
Cross-Level Visual--Geometric Interaction (CVGI) and
Question-Conditioned Historical Evidence Re-reading
(QHER). The frozen Qwen visual encoder~\cite{bai2025qwen25vl}
and causal StreamVGGT geometry encoder
~\cite{zhuo2026streamvggt} provide complementary
representations. During observation, CVGI injects
multilevel geometry into evolving visual representations,
while Historical Evidence Memory $\mathcal M_t^E$
preserves bounded, source-indexed visual--geometric
features. At answer time, QHER constructs source-bound
Evidence Tokens and retrieves corresponding historical
geometry through multilevel Geometry Readers.
Separately, Bounded Dual-KV Management (BDKM) serves
as a training-free inference extension, independently
managing Geometry KV $\mathcal C_t^G$ and Qwen KV
$\mathcal C_t^Q$ under separate capacity budgets, with
representative anchors protected only in Qwen KV.
Historical Evidence Memory remains independent of
both runtime KV caches.

\subsection{Cross-Level Visual--Geometric Interaction}
\label{sec:method_geometry}

\noindent\textbf{Causal feature extraction and alignment.}
For each incoming observation $I_t$, the frozen Qwen
visual encoder and merger produce
$V_t^0=F_{\mathrm{vis}}(I_t)$. In parallel, StreamVGGT
processes the observation using its causal KV state
and extracts geometric features at three levels
$\mathcal L_G=\{11,17,23\}$:
\begin{equation}
    \bigl(\{G_t^{(\ell)}\}_{\ell\in\mathcal L_G},
    \mathcal C_t^G\bigr)
    =F_{\mathrm{geo}}(I_t,\mathcal C_{t-1}^G).
    \label{eq:causal_geometry}
\end{equation}
Spatial patch features support visual--geometric
interaction, while camera and register tokens participate
in geometry encoding. To align geometry with Qwen visual
tokens, spatially aligned $2\times2$ geometry patches
are concatenated along the channel dimension and mapped
into Qwen's hidden space through a shared trainable
projector:
\begin{equation}
    Z_t^{(\ell)}
    =P\!\left(
    \operatorname{Align}_{2\times2}(G_t^{(\ell)})
    \right),
    \quad \ell\in\mathcal L_G.
    \label{eq:geometry_alignment}
\end{equation}

\noindent\textbf{Observation-time visual--geometric fusion.}
CVGI first augments the input visual embeddings with
a gated residual from the final geometry level $G^{23}$:
\begin{equation}
    \widetilde V_t^0
    =V_t^0+\tanh(a)\,Z_t^{(23)}.
    \label{eq:online_input_geometry}
\end{equation}
It then injects $G^{11}$, $G^{17}$, and $G^{23}$
after Qwen blocks $\mathcal K_A=\{0,1,2\}$,
respectively. All block indices use zero-based indexing.
Let $H_t^{(k)}$ denote the visual hidden states immediately
after block $k$ and before geometric interaction:
\begin{equation}
\begin{aligned}
    \widetilde H_t^{(k)}
    =H_t^{(k)}
    +\tanh(\alpha_k)\,
    \sigma\!\left(g_k(H_t^{(k)})\right)
    \odot Z_t^{(\ell(k))},
\end{aligned}
    \label{eq:online_layer_geometry}
\end{equation}
where $\ell(0)=11$, $\ell(1)=17$, and $\ell(2)=23$.
Here, $g_k$ predicts a visual-token-level gate, while
$\alpha_k$ controls the overall residual strength.
The input residual parameter $a$ and layer scales
$\alpha_k$ are initialized to zero; gate logits are
also initialized to zero, yielding an initial sigmoid
value of $0.5$. Thus, geometric residual updates start
at zero, while all associated parameters remain trainable.

In parallel, Historical Evidence Memory $\mathcal M_t^E$
retains pre-fusion visual features and spatially aligned,
pre-projection geometric patches from completed observations,
together with their source correspondence, for subsequent
QHER retrieval.

\subsection{Question-Conditioned Historical Evidence Re-reading}
\label{sec:method_qher}
\label{sec:method_memory}
\label{sec:method_requery}
\label{sec:method_reading}

\noindent\textbf{Question-agnostic historical evidence preservation.}
QHER separates question-agnostic evidence preservation
from question-conditioned retrieval at answer time.
For each completed observation $f$,
$\mathcal M_t^E$ stores pre-fusion visual features,
aligned multilevel geometric features before projection,
spatial coordinates, validity masks, and source metadata.
It retains up to eight recent and eight historical
observations, with historical entries compressed into
grids of at most $7\times7$, preserving spatial
correspondence between visual regions and geometry
independently of future questions.

\noindent\textbf{Question-conditioned evidence construction.}
When a question $q_\tau$ arrives, its representation
$h_\tau^q$ is extracted from the final normalized Qwen
hidden state at the last valid question token and retained
during subsequent observations. At answer time $t_s$,
it is combined with the latest visual summary:
\begin{equation}
    c_\tau=\phi_q\!\left(
        h_\tau^q,\operatorname{Pool}(V_{t_s}^0)
    \right).
    \label{eq:question_condition}
\end{equation}
This condition guides patch- and frame-level relevance
estimation using visual relevance, pooled context, and
temporal distance:
\begin{equation}
\begin{aligned}
    s_f &=
    \operatorname{Score}_{\mathrm{frame}}
    (c_\tau,\mathbf w_f,\bar V_f,\Delta t_f),\\
    \pi_f &= \operatorname{Softmax}_{f\in\mathcal F_t}(s_f),
\end{aligned}
    \label{eq:frame_priority}
\end{equation}
where $\mathbf w_f=\{w_{f,j}\}_j$ and
$\bar V_f=\operatorname{Pool}(V_f^0)$.
The priorities provide soft relevance estimates rather
than hard frame selection: tokens are constructed for
every nonempty region, while $\pi_f$ modulates subsequent
geometric write gates.

Each retained observation is divided into four spatial
regions $\mathcal R_{f,r}$. For each nonempty region,
QHER constructs a question-weighted local token and
a mean-pooled context token:
\begin{equation}
\begin{aligned}
    z_{f,r}^{\mathrm{loc}}
    &=
    \frac{\sum_{j\in\mathcal R_{f,r}}
    w_{f,j}V_{f,j}^0}
    {\sum_{j\in\mathcal R_{f,r}}w_{f,j}+\epsilon},\\
    z_{f,r}^{\mathrm{ctx}}
    &=
    \frac{1}{|\mathcal R_{f,r}|}
    \sum_{j\in\mathcal R_{f,r}}V_{f,j}^0.
\end{aligned}
    \label{eq:evidence_representation}
\end{equation}
Token-type, region, and relative-time embeddings are
then added:
\begin{equation}
\begin{aligned}
    e_{f,r}^{s}
    &=z_{f,r}^{s}+E_{\mathrm{type}}(s)\\
    &\quad+E_{\mathrm{region}}(r)
    +E_{\mathrm{time}}(\Delta t_f),\\
    &\qquad s\in\{\mathrm{loc},\mathrm{ctx}\}.
\end{aligned}
    \label{eq:evidence_embedding}
\end{equation}
Each token retains its source-observation address for
geometry retrieval. With four regions and at most two
tokens per region, each observation contributes at most
eight tokens, yielding at most 128 tokens for sixteen
retained observations. They are appended after the
completed user message and before the assistant header,
without re-encoding historical frames.


\begin{table*}[t]
  \centering
  \caption{Stream3D-Bench results at 1 FPS. NA, MCA, and
OEA denote numerical, multiple-choice, and open-ended QA;
ATA denotes answer-timing accuracy. Published results
are reported from the respective papers; FT denotes
fine-tuned. Best and second-best values per metric are
bold and underlined, respectively.}
  \label{tab:stream3d_main}
  \begingroup
  \vigerMainTableStyle
  \begin{tabularx}{\vigerMainTableWidth}
    {@{}l|c|ZZZ|ZZZ|ZZZ|c c c c c@{}}

    \toprule
    \multicolumn{1}{c|}{\multirow[c]{2}{*}{\textbf{Method}}} &
    \multirow[c]{2}{*}{\textbf{Avg.}$\uparrow$} &
    \multicolumn{3}{c|}{\textbf{Backward Tracing}} &
    \multicolumn{3}{c|}{\textbf{Realtime Perception}} &
    \multicolumn{3}{c|}{\textbf{Forward Response}} &
    \textbf{Answer-} &
    \textbf{TTFT} &
    \textbf{End2End} &
    \textbf{Memory} &
    \textbf{Image} \\

    \cmidrule(lr){3-5}
    \cmidrule(lr){6-8}
    \cmidrule(lr){9-11}

    & &
    NA & MCA & OEA &
    NA & MCA & OEA &
    NA & MCA & OEA &
    \textbf{Timing Acc.}$\uparrow$ &
    \textbf{Latency}$\downarrow$ &
    \textbf{Latency}$\downarrow$ &
    \textbf{Usage}$\downarrow$ &
    \textbf{Resolution} \\
    \midrule

    \rowcolor{vigerTableGroup}
    \multicolumn{16}{@{}l}{\emph{Proprietary Models (API)}} \\

    GPT-4o &
    28.0 &
    10.5 &
    36.3 &
    24.9 &
    31.1 &
    34.8 &
    51.1 &
    10.2 &
    29.9 &
    23.0 &
    55.9\% &
    -- &
    -- &
    -- &
    1296$\times$968 \\

    GPT-5 &
    35.0 &
    18.7 &
    44.1 &
    33.6 &
    38.0 &
    46.2 &
    52.8 &
    12.9 &
    30.6 &
    37.7 &
    61.7\% &
    -- &
    -- &
    -- &
    1296$\times$968 \\

    \midrule

    \rowcolor{vigerTableGroup}
    \multicolumn{16}{@{}l}{\emph{Open-source Models}} \\

    LLaVA-Video-7B &
    20.4 &
    12.7 &
    34.6 &
    11.4 &
    21.6 &
    28.6 &
    38.0 &
    10.2 &
    20.3 &
    6.5 &
    55.9\% &
    650~ms &
    2.84~s &
    25.8~G &
    1152$\times$768 \\

    InternVL3-8B &
    24.1 &
    22.1 &
    34.4 &
    6.8 &
    30.7 &
    31.2 &
    40.0 &
    18.3 &
    27.2 &
    5.9 &
    19.8\% &
    1034~ms &
    5.65~s &
    40.5~G &
    1024$\times$765 \\

    InternVL3.5-8B &
    27.0 &
    21.4 &
    36.2 &
    17.6 &
    28.2 &
    38.2 &
    42.4 &
    25.2 &
    25.8 &
    8.2 &
    23.0\% &
    1771~ms &
    6.72~s &
    43.1~G &
    1024$\times$765 \\

    Qwen2.5-VL-7B &
    22.5 &
    11.6 &
    36.3 &
    13.6 &
    23.1 &
    33.2 &
    39.4 &
    11.3 &
    28.4 &
    5.5 &
    19.2\% &
    325~ms &
    3.16~s &
    40.0~G &
    1008$\times$784 \\

    Qwen2.5-VL-32B &
    29.3 &
    14.1 &
    40.3 &
    35.7 &
    28.7 &
    37.6 &
    43.2 &
    22.4 &
    30.2 &
    11.5 &
    35.2\% &
    560~ms &
    5.87~s &
    91.6~G &
    1008$\times$784 \\

    Qwen3-VL-8B &
    19.0 &
    2.5 &
    33.2 &
    20.2 &
    16.0 &
    33.4 &
    28.4 &
    4.6 &
    24.7 &
    8.3 &
    61.7\% &
    150~ms &
    1.56~s &
    28.8~G &
    1024$\times$768 \\

    Qwen3-VL-32B &
    23.1 &
    4.9 &
    34.2 &
    35.9 &
    19.4 &
    32.0 &
    31.8 &
    7.3 &
    29.2 &
    13.5 &
    66.4\% &
    282~ms &
    3.22~s &
    65.5~G &
    1024$\times$768 \\

    \midrule

    \rowcolor{vigerTableGroup}
    \multicolumn{16}{@{}l}{\emph{Streaming Dialogue VLMs}} \\

    VideoLLM-online-8B (FT) &
    34.6 &
    28.8 &
    40.9 &
    27.8 &
    31.3 &
    37.9 &
    44.6 &
    38.2 &
    32.7 &
    29.2 &
    70.2\% &
    120~ms &
    1.29~s &
    28.4~G &
    384$\times$384 \\

    Qwen2.5-VL-7B (FT) &
    47.8 &
    45.6 &
    56.6 &
    45.4 &
    45.7 &
    44.6 &
    48.7 &
    46.6 &
    57.4 &
    39.8 &
    73.1\% &
    106~ms &
    0.53~s &
    36.8~G &
    504$\times$392 \\

    Stream3D-VLM-4B &
    54.6 &
    58.1 &
    61.2 &
    45.2 &
    58.8 &
    57.3 &
    52.8 &
    55.3 &
    60.6 &
    41.9 &
    75.4\% &
    \textbf{43~ms} &
    \textbf{0.24~s} &
    \underline{20.7~G} &
    504$\times$392 \\

    Stream3D-VLM-8B &
    \underline{58.8} &
    \underline{59.8} &
    \underline{67.9} &
    \underline{50.5} &
    \underline{61.4} &
    \underline{61.5} &
    \underline{54.3} &
    \underline{58.1} &
    \underline{66.7} &
    \underline{49.4} &
    \underline{86.7\%} &
    \underline{62~ms} &
    0.39~s &
    36.6~G &
    504$\times$392 \\

    \midrule

    \rowcolor{vigerTableOurs}
    \textbf{GeoRecall-4B (Ours)} &
    \textbf{71.4} &
    \textbf{66.1} &
    \textbf{84.1} &
    \textbf{57.5} &
    \textbf{65.7} &
    \textbf{80.7} &
    \textbf{58.8} &
    \textbf{73.0} &
    \textbf{90.6} &
    \textbf{66.4} &
    \textbf{96.5\%} &
    124.0~ms &
    \underline{0.35~s} &
    \textbf{15.0~G} &
    504$\times$392 \\

    \bottomrule

  \end{tabularx}
  \endgroup
\end{table*}

\begin{table*}[t]
  \centering
  \caption{VSI-Bench results. Numerical tasks use mean
  relative accuracy, while multiple-choice tasks
  use accuracy. Published task scores are taken from
the cited evaluation sources. Best and second-best scores
  are bold and underlined, respectively.}
  \label{tab:vsi_main}
  \begingroup
  \vigerMainTableStyle
  \begin{tabularx}{0.92\textwidth}
    {@{\hspace{5pt}}l|c|c|ZZZZ|ZZZZ@{\hspace{5pt}}}

    \toprule
    \multicolumn{1}{c|}{\multirow[c]{2}{*}{\textbf{Method}}} &
    \multirow[c]{2}{*}{\textbf{Online}} &
    \multirow[c]{2}{*}{\textbf{Avg.}$\uparrow$} &
    \multicolumn{4}{c|}{\textbf{Numerical Answer}} &
    \multicolumn{4}{c}{\textbf{Multiple-Choice Answer}} \\

    \cmidrule(lr){4-7}
    \cmidrule(lr){8-11}

    & & &
    \mbox{Obj. Cnt.} &
    \mbox{Abs. Dist.} &
    \mbox{Obj. Size} &
    \mbox{Room Size} &
    \mbox{Rel. Dist.} &
    \mbox{Rel. Dir.} &
    \mbox{Route Plan} &
    \mbox{Appr. Order} \\
    \midrule

    \rowcolor{vigerTableGroup}
    \multicolumn{11}{@{}l}{\emph{Proprietary Models (API)}} \\

    GPT-4o &
    \vigerNo &
    34.0 &
    46.2 &
    5.3 &
    43.8 &
    38.2 &
    37.0 &
    41.3 &
    31.5 &
    28.5 \\

    Gemini-1.5 Pro &
    \vigerNo &
    45.4 &
    56.2 &
    30.9 &
    64.1 &
    43.6 &
    51.3 &
    46.3 &
    36.0 &
    34.6 \\

    Gemini-2.5 Pro &
    \vigerNo &
    51.5 &
    43.8 &
    34.9 &
    64.3 &
    42.8 &
    61.1 &
    47.8 &
    \underline{45.9} &
    71.3 \\
    \midrule

    \rowcolor{vigerTableGroup}
    \multicolumn{11}{@{}l}{\emph{Open-source Models}} \\

    LongVILA-8B &
    \vigerNo &
    21.7 &
    29.1 &
    9.1 &
    16.7 &
    0.0 &
    29.6 &
    30.7 &
    32.5 &
    25.5 \\

    VILA-1.5-40B &
    \vigerNo &
    31.2 &
    22.4 &
    24.8 &
    48.7 &
    22.7 &
    40.5 &
    25.7 &
    31.5 &
    32.9 \\

    Qwen2.5-VL-7B &
    \vigerNo &
    31.4 &
    40.9 &
    14.8 &
    43.4 &
    10.7 &
    38.6 &
    40.1 &
    33.0 &
    29.8 \\

    Qwen2.5-VL-72B &
    \vigerNo &
    34.4 &
    25.1 &
    29.3 &
    57.9 &
    29.4 &
    41.7 &
    39.3 &
    23.2 &
    29.0 \\

    LLaVA-OneVision-72B &
    \vigerNo &
    40.3 &
    43.5 &
    23.9 &
    57.6 &
    37.5 &
    42.5 &
    39.9 &
    32.5 &
    44.6 \\

    LLaVA-NeXT-Video-72B &
    \vigerNo &
    40.9 &
    48.9 &
    22.8 &
    57.4 &
    35.3 &
    42.4 &
    36.7 &
    35.0 &
    48.6 \\
    \midrule

    \rowcolor{vigerTableGroup}
    \multicolumn{11}{@{}l}{\emph{Spatial Reasoning Models}} \\

    SpaceR-7B &
    \vigerNo &
    45.5 &
    57.8 &
    28.2 &
    59.9 &
    47.1 &
    40.1 &
    45.4 &
    33.5 &
    52.1 \\

    Spatial-MLLM-4B &
    \vigerNo &
    47.0 &
    65.3 &
    34.8 &
    63.1 &
    45.1 &
    41.3 &
    46.9 &
    33.5 &
    46.3 \\

    VG LLM-8B &
    \vigerNo &
    50.7 &
    67.9 &
    37.7 &
    58.6 &
    62.0 &
    46.6 &
    40.7 &
    32.4 &
    59.2 \\

    VLM-3R-7B &
    \vigerNo &
    60.9 &
    \underline{70.2} &
    \underline{49.4} &
    69.2 &
    \underline{67.1} &
    \underline{65.4} &
    \textbf{80.5} &
    45.4 &
    40.1 \\

    Stream3D-VLM-4B &
    \vigerYes &
    55.2 &
    68.6 &
    38.4 &
    65.9 &
    56.4 &
    49.9 &
    54.3 &
    42.3 &
    66.0 \\

    Stream3D-VLM-8B &
    \vigerYes &
    \textbf{65.9} &
    \textbf{72.4} &
    \textbf{50.1} &
    \textbf{71.5} &
    \textbf{68.7} &
    \textbf{70.8} &
    \underline{73.3} &
    \textbf{46.2} &
    \underline{73.8} \\
    \midrule

    \rowcolor{vigerTableOurs}
    \textbf{GeoRecall-4B (Ours)} &
    \vigerYes &
    \underline{62.0} &
    70.1 &
    42.7 &
    \underline{69.3} &
    60.1 &
    61.3 &
    69.2 &
    39.4 &
    \textbf{83.5} \\
    \bottomrule

  \end{tabularx}
  \endgroup
\end{table*}

\noindent\textbf{Source-constrained progressive geometry reading.}
For each Evidence Token $e$, Geometry Readers retrieve
features only from its associated source observation
$f(e)$. They operate after Qwen blocks
$\mathcal K_B=\{1,3,5\}$, paired with $G^{11}$,
$G^{17}$, and $G^{23}$, respectively (zero-based
indexing). Unlike CVGI, they update Evidence Token
states rather than current visual representations.
A question-conditioned query is formed as:
\begin{equation}
\begin{aligned}
    Q_e^{(k)}
    &=W_Q^{(k)}\operatorname{LN}(h_e^{(k)})\\
    &\quad+\rho_k W_c^{(k)}
    \operatorname{LN}(c_\tau).
\end{aligned}
    \label{eq:reader_query}
\end{equation}
Geometric keys and values are projected from the source
observation, with spatial coordinates incorporated
into the keys. Source-constrained attention combines
a validity mask and spatial prior:
\begin{equation}
\begin{aligned}
    S_e^{(k)}
    &=\frac{Q_e^{(k)}(K_{f(e)}^{(k)})^\top}
    {\sqrt{d_k}},\\
    A_e^{(k)}
    &=\operatorname{Softmax}
    \left(S_e^{(k)}+B_e^{(k)}+M_{f(e)}\right).
\end{aligned}
    \label{eq:geometry_attention}
\end{equation}
Frame priority further modulates the geometric write gate:
\begin{equation}
\begin{aligned}
    u_{e,k}
    &=g_k^{\mathrm{write}}
    (h_e^{(k)},c_\tau,E_{\mathrm{type}}(e))\\
    &\quad+\log(N\pi_{f(e)}+\epsilon),\\
    \beta_{e,k}&=\sigma(u_{e,k}),
\end{aligned}
    \label{eq:geometry_write_gate}
\end{equation}
where $N=|\mathcal F_t|$ is the number of retained
observations. Retrieved geometry is projected into
Qwen's hidden space and integrated through a gated
residual:
\begin{equation}
\begin{aligned}
    r_e^{(k)}
    &=W_O^{(k)}
    \left(A_e^{(k)}U_{f(e)}^{(k)}\right),\\
    \widetilde h_e^{(k)}
    &=h_e^{(k)}+\gamma_k\beta_{e,k}r_e^{(k)}.
\end{aligned}
    \label{eq:geometry_readout}
\end{equation}
The Reader output projection is zero-initialized,
while $\gamma_k$ is initialized to $0.1$.

Within each source observation, local tokens use a
smoothed spatial prior favoring their corresponding
regions while allowing access to other valid locations;
context tokens attend to all valid locations. Successive
Readers and intervening Qwen blocks progressively refine
historical evidence according to the current question,
without re-encoding past frames.

\subsection{Bounded Dual-KV Management}
\label{sec:method_cache}

For long-stream inference, BDKM provides auxiliary,
training-free cache management with independent capacity
budgets for StreamVGGT Geometry KV and Qwen KV.
It applies attention-based token pruning to Geometry KV
and frame-group eviction to Qwen KV. Representative
anchors selected using joint visual--geometric features
receive additional protection only in Qwen KV, within
its fixed budget. Both caches remain separate from
the Historical Evidence Memory used by QHER.
Details of the cache policies and configuration are
provided in Appendix~C.1.

\begin{table*}[t]
  \centering
  \caption{Comparison on OST-Bench and MindCube-Tiny
  with representative models up to 7B parameters.
  $\dagger$ denotes our independently retrained
  Stream3D-VLM-4B baseline. Best and second-best
  results are bold and underlined, respectively.}
  \label{tab:ost_mindcube_main}
  \begingroup
  \vigerMainTableStyle
  \begin{tabularx}{0.92\textwidth}
    {@{\hspace{7pt}}l@{\hspace{8pt}}|@{\hspace{7pt}}c|ZZZZ|ZZZZ@{\hspace{5pt}}}

    \toprule
    \multicolumn{1}{c|}{\multirow[c]{2}{*}{\textbf{Method}}} &
    \multirow[c]{2}{*}{\textbf{Size}} &
    \multicolumn{4}{c|}{\textbf{OST-Bench}} &
    \multicolumn{4}{c}{\textbf{MindCube-Tiny}} \\

    \cmidrule(lr){3-6}
    \cmidrule(lr){7-10}

    & &
    \textbf{Avg.}$\uparrow$ &
    \mbox{State} &
    \mbox{Info} &
    \mbox{AO} &
    \textbf{Avg.}$\uparrow$ &
    \mbox{Rot.} &
    \mbox{Among} &
    \mbox{Around} \\
    \midrule

    \rowcolor{vigerTableGroup}
    \multicolumn{10}{@{\hspace{7pt}}l}{\emph{Offline Spatial MLLMs}} \\

    Cambrian-S-7B &
    7B &
    41.3 &
    37.2 &
    \textbf{58.5} &
    32.1 &
    37.8 &
    33.0 &
    38.8 &
    39.2 \\

    VST-7B &
    7B &
    \textbf{43.7} &
    \textbf{43.0} &
    \underline{58.1} &
    \underline{33.5} &
    39.0 &
    36.5 &
    37.8 &
    44.0 \\

    Spatial-MLLM-4B &
    4B &
    15.7 &
    24.3 &
    23.6 &
    12.7 &
    33.0 &
    \underline{40.0} &
    29.2 &
    36.8 \\

    VLM-3R-7B &
    7B &
    \underline{43.3} &
    \underline{39.0} &
    \textbf{58.5} &
    \textbf{35.4} &
    38.6 &
    33.5 &
    43.3 &
    31.2 \\

    G$^2$VLM-2B &
    2B &
    16.6 &
    24.1 &
    25.9 &
    12.9 &
    41.0 &
    36.0 &
    \underline{44.3} &
    36.8 \\

    Cambrian-P-7B &
    7B &
    39.6 &
    36.6 &
    56.7 &
    31.0 &
    35.1 &
    37.0 &
    33.2 &
    38.4 \\

    \midrule

    \rowcolor{vigerTableGroup}
    \multicolumn{10}{@{\hspace{7pt}}l}{\emph{Streaming Spatial MLLMs}} \\

    Spatial-TTT-nano &
    2B &
    28.4 &
    26.1 &
    39.4 &
    24.3 &
    \textbf{71.3} &
    \textbf{58.0} &
    \textbf{70.7} &
    \textbf{83.6} \\

    Stream3D-VLM-4B$^{\dagger}$ &
    4B &
    19.9 &
    27.1 &
    28.0 &
    16.8 &
    33.1 &
    27.5 &
    30.0 &
    44.8 \\

    \rowcolor{vigerTableOurs}
    \textbf{GeoRecall-4B (Ours)} &
    4B &
    30.6 &
    31.3 &
    38.6 &
    26.2 &
    \underline{45.0} &
    37.5 &
    \underline{44.3} &
    \underline{52.8} \\

    \bottomrule

  \end{tabularx}
  \endgroup
\end{table*}

\section{Experiments}
\label{sec:experiments}

\subsection{Experimental Setup}
\label{sec:exp_setup}

\subsubsection{Datasets}

We evaluate GeoRecall on four benchmarks:
Stream3D-Bench~\cite{yu2026stream3dvlm},
VSI-Bench~\cite{yang2025vsibench},
OST-Bench~\cite{lin2025ostbench}, and
MindCube-Tiny~\cite{wang2026mindcube}.
These benchmarks cover streaming spatial QA,
spatial reasoning, online spatio-temporal understanding,
and multi-view spatial reasoning.

\subsubsection{Evaluation Metrics}

For Stream3D-Bench, we report overall QA scores,
nine subtask scores, and answer-timing accuracy (ATA).
Stream3D-VLM and GeoRecall share the same item-level
open-ended answer (OEA) scoring protocol; other reference
results retain their published evaluation protocols.
For VSI-Bench, numerical and multiple-choice tasks
are evaluated using mean relative accuracy and accuracy,
respectively. For OST-Bench and MindCube-Tiny, we report
task-wise and average scores following their public
evaluation procedures. Inference latency and GPU memory
usage are reported in the corresponding experiments.
Appendix~B.4 details the OEA scoring rules and resource
measurement conventions.

\subsubsection{Comparison Methods}

We compare against four groups of methods:
proprietary API models, including
GPT-4o~\cite{openai2024gpt4o},
GPT-5~\cite{openai2025gpt5},
Gemini-1.5 Pro~\cite{geminiteam2024gemini15},
and Gemini-2.5 Pro~\cite{comanici2025gemini25};
general-purpose VLMs, including
LLaVA-Video~\cite{zhang2025llavavideo},
InternVL3~\cite{zhu2025internvl3},
InternVL3.5~\cite{wang2025internvl35},
Qwen2.5-VL~\cite{bai2025qwen25vl},
Qwen3-VL~\cite{bai2025qwen3vl},
LongVILA~\cite{chen2025longvila},
VILA-1.5~\cite{lin2024vila},
LLaVA-OneVision~\cite{li2024llavaonevision},
and LLaVA-NeXT-Video~\cite{zhang2024llavanextvideo};
streaming VLMs, including
VideoLLM-online~\cite{chen2024videollmonline},
Spatial-TTT-nano~\cite{liu2026spatialttt},
and Stream3D-VLM~\cite{yu2026stream3dvlm};
and offline spatial reasoning models, including
SpaceR~\cite{ouyang2025spacer},
Spatial-MLLM~\cite{wu2025spatialmllm},
VG-LLM~\cite{zheng2025vgllm},
VLM-3R~\cite{fan2026vlm3r},
VST~\cite{yang2025vst},
G$^2$VLM~\cite{hu2026g2vlm},
Cambrian-S~\cite{yang2026cambrians},
and Cambrian-P~\cite{yang2026cambrianp}.

\subsubsection{Implementation Details}

GeoRecall is built on Qwen2.5-VL-3B-Instruct
and a pretrained causal StreamVGGT geometry encoder.
The visual encoder, visual merger, and geometry encoder
are frozen, while the language-model backbone, geometric
projector, and CVGI/QHER modules are trainable.
Training uses BF16 precision on eight NVIDIA H200 GPUs.
Stream3D-Bench and VSI-Bench use greedy decoding, while
OST-Bench follows its official per-round evaluation protocol.
MindCube-Tiny uses option-log-probability aggregation
across cyclic permutations of answer options. Additional
implementation details are provided in Appendix B.

\begin{figure*}[t]
    \centering
    \IfFileExists{figures/compare.pdf}{%
        \includegraphics[width=0.96\textwidth]
        {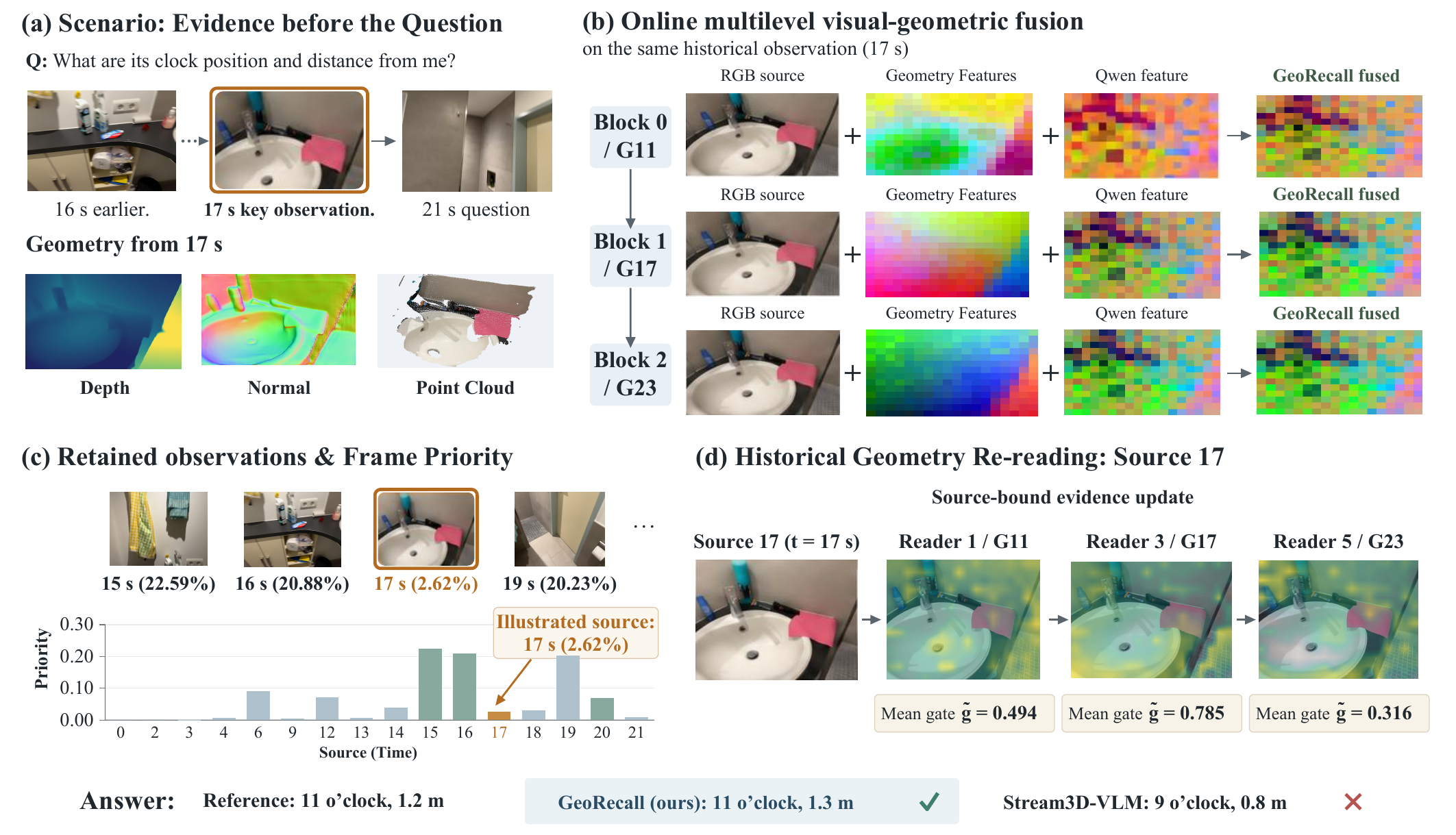}%
    }{%
        \fbox{\parbox[c][1.15in][c]{0.93\textwidth}
        {\centering Qualitative comparison placeholder.}}%
    }
    \caption{
    Qualitative illustration of GeoRecall.
    (a) Evidence precedes the question.
    (b) Online multilevel visual--geometric fusion.
    (c) Question-conditioned frame priorities.
    (d) Multilevel geometric re-reading from the same source.
    Final predictions are compared with the reference answer and Stream3D-VLM.
    }
    \label{fig:qualitative_comparison}
\end{figure*}

\subsection{Main Results}
\label{sec:exp_main}

Table~\ref{tab:stream3d_main} reports the results on
Stream3D-Bench. GeoRecall-4B achieves an overall QA score
of $71.4$, exceeding Stream3D-VLM-8B ($58.8$) by
$12.6$ points. Its MCA gains extend across Backward
Tracing, Realtime Perception, and Forward Response,
indicating improved answer accuracy under different
temporal relationships between questions and observations.
These settings require both incorporating newly arriving
spatial information and accessing previously observed
evidence. This requirement aligns with the roles of CVGI
and QHER: CVGI integrates geometry during observation,
while QHER provides question-conditioned access to
retained geometry when an answer is generated.
Additional qualitative examples on Stream3D-Bench
are provided in Appendix~D.

On VSI-Bench (Table~\ref{tab:vsi_main}), GeoRecall-4B
achieves $62.0$, improving over Stream3D-VLM-4B
($55.2$) by $6.8$ points. The gains are particularly
pronounced in Relative Distance ($+11.4$), Relative
Direction ($+14.9$), and Appearance Order ($+17.5$).
The first two tasks require reasoning about relationships
between objects, while the third requires distinguishing
their observations over time. This pattern is consistent
with combining aligned geometric representations and
source-indexed historical evidence. However, Route Planning
decreases from $42.3$ to $39.4$, suggesting that improved
spatial evidence access does not automatically translate
into stronger multi-step route reasoning. 

Table~\ref{tab:ost_mindcube_main} extends the evaluation
to online spatio-temporal understanding and multi-view
spatial reasoning. On OST-Bench, GeoRecall-4B obtains
$30.6$, compared with $28.4$ for Spatial-TTT-nano
and $19.9$ for Stream3D-VLM-4B, improving over the
latter across State, Info, and AO. On MindCube-Tiny,
it achieves $45.0$, exceeding the compared offline
spatial VLMs and Stream3D-VLM-4B ($33.1$), while
remaining below Spatial-TTT-nano ($71.3$).
Its Among and Around scores ($44.3$ and $52.8$)
exceed its Rotation score ($37.5$), showing that
performance remains uneven across spatial tasks.
These evaluations complement the streaming results.

\subsection{Qualitative Analysis}
\label{sec:exp_qualitative}

Fig.~\ref{fig:qualitative_comparison} illustrates how
GeoRecall handles spatial evidence that precedes a
question. As shown in (a), the key observation appears
at 17 s, whereas the question arrives at 21 s from a
different view, making historical evidence necessary.
During observation, CVGI progressively integrates
multilevel geometric features into visual representations
at successive language-model blocks, while retaining
source-aligned visual--geometric evidence for later
access (b). After the question arrives, QHER evaluates
the relevance of retained observations according to
the current query (c). Importantly, low frame priority
does not exclude a retained source, but instead
modulates the Reader write gates. The Geometry Readers
then progressively access multilevel geometry from
the corresponding historical source to update its
Evidence Tokens (d). In this example, GeoRecall
provides more accurate estimates of the target's
direction and distance than Stream3D-VLM, producing
predictions closer to the reference. This illustrates
the complementary roles of CVGI and QHER: online
visual--geometric integration during observation and
question-conditioned retrieval of historical geometry
at answer time enable effective reasoning over spatial
evidence that is no longer visible in the current view.

\subsection{Architecture Comparison and Component Ablations}
\label{sec:exp_ablation}

Table~\ref{tab:architecture_ablation} compares full-benchmark
architectures (Group A) and controlled ablations (Group B);
Group A contrasts GeoRecall with alternatives. Multi-Level
Routing (A2) predicts geometry-layer mixing weights and residual
strengths from a learnable See-Query, injecting mixed features
through attention-based Readers. GeoRecall instead establishes
layer-wise correspondence between geometric levels and Qwen
blocks, using visually conditioned gated residual injection.
Question-Conditioned Routing (A3) uses question representations
to guide geometric routing during online observation, whereas
GeoRecall keeps online fusion question-agnostic and applies
question conditioning only to historical reading at answer time.
Historical Reading (A4) retrieves mixed geometry from a historical
window through Readers at blocks 4, 15, and 24. GeoRecall instead
applies CVGI at blocks 0, 1, and 2 and QHER at blocks 1, 3, and 5,
constraining each Evidence Token to geometry from its
corresponding source observation.

Group B evaluates component ablations using models trained
for 8,000 steps under matched training settings. All variants
are evaluated on the same fixed subset of 500 questions,
comprising 300 Stream3D-Bench and 200 VSI-Bench questions.
The ablations remove online fusion (B2), replace multilevel
online geometry with $G^{23}$ alone (B4), remove historical
Evidence Tokens and Readers (B3), or replace explicit
question conditioning in the reading controller with a
shared trainable vector (B5). These ablations examine the
roles of online visual--geometric interaction, multilevel
geometric diversity, historical evidence retrieval, and
question-guided reading. The results indicate that online
geometry integration and historical evidence access are
complementary, while multilevel representations and explicit
question conditioning further support streaming spatial
reasoning.

\begin{table}[t]
\centering
\caption{Architecture comparison and component ablations
of GeoRecall. Group A compares full-benchmark architectures;
Group B evaluates controlled ablations on 500 questions
(300 Stream3D-Bench and 200 VSI-Bench). All is the
arithmetic mean of the two benchmark scores.}
\label{tab:architecture_ablation}

\scriptsize
\setlength{\tabcolsep}{2.2pt}
\renewcommand{\arraystretch}{1.08}

\begin{tabular}{@{}c
                >{\raggedright\arraybackslash}p{0.39\columnwidth}
                r
                c
                c
                c@{}}
\toprule
ID & Method & Steps & All$\uparrow$ & Stream3D-Bench$\uparrow$ & VSI-Bench$\uparrow$\\
\midrule

\rowcolor{vigerTableGroup}
\multicolumn{6}{@{}l}{\emph{A. Historical Architecture Comparison}}\\

A1 & GeoRecall & 49,801 & \textbf{66.7} & \textbf{71.4} & \textbf{62.0}\\
A2 & Multi-Level Routing & 49,801 & 62.2 & 65.8 & 58.6\\
A3 & Question-Cond. Routing & 49,801 & 61.1 & 64.9 & 57.2\\
A4 & Historical Reading (4/15/24) & 49,801 & 60.5 & 62.2 & 58.7\\

\midrule

\rowcolor{vigerTableGroup}
\multicolumn{6}{@{}l}{\emph{B. Component Ablation}}\\

B1 & GeoRecall  & 8,000 & \textbf{54.8} & \textbf{57.0} & 52.5 \\
B2 & w/o Online Fusion & 8,000 & 48.3 & 46.9 & 49.7\\
B3 & w/o Historical Reading & 8,000 & 51.4 & 52.6 & 50.2\\
B4 & G23-only & 8,000 & 48.9 & 46.7 & 51.1\\
B5 & w/o Explicit Query Conditioning & 8,000 & 51.7 & 50.5 & \textbf{52.8}\\

\bottomrule
\end{tabular}
\end{table}

\subsection{BDKM as a Training-Free Inference Extension}
\label{sec:bdkm_extension}

We evaluate BDKM through cache-policy diagnostics,
resource measurements, and long-video QA. Cache-policy
diagnostics identify joint visual--geometric anchor
protection in Qwen KV as the preferred configuration
among policies that bound both caches. In a streaming
stress test with 5,000 distinct observations uniformly
sampled from a single video, Full Cache runs out of
memory at observation 794, whereas BDKM completes
the entire sequence with a peak allocated memory of
19.9 GiB by independently bounding Geometry KV and
Qwen KV. We further compare 32-frame sampling with
native frame-wise processing on 80 randomly selected
questions from 30 videos, grouped into three video-length
ranges: 1,000--3,499, 3,500--6,499, and 6,500--12,000
frames. While frame-wise processing improves QA by
9.26 points in the intermediate-length group, it
decreases by 13.70 and 25.00 points in the shorter and
longest groups, respectively, resulting in lower overall
performance than 32-frame sampling. This performance
variation may be related to the mismatch between shorter
training sequences and longer inference streams, although
differences in question placement and processing
granularity may also contribute. These results show
that BDKM enables bounded long-stream inference, while
maintaining effective spatial reasoning over extended
videos remains challenging. Further analysis is provided in Appendix~C.1 and C.2.
\section{Conclusion}
\label{sec:conclusion}

GeoRecall addresses streaming spatial question answering
by combining online multilevel visual--geometric
integration with question-conditioned historical evidence
retrieval. CVGI injects multilevel geometric features into
evolving visual representations during observation,
while QHER constructs source-bound Evidence Tokens and
retrieves their corresponding historical geometry at
answer time, without re-encoding past frames. As a
separately evaluated, training-free inference extension,
BDKM independently manages Geometry and Qwen KV caches
under separate capacity budgets to support bounded-memory
streaming inference. Experiments across four benchmarks
show that GeoRecall achieves an overall QA score of 71.4
on Stream3D-Bench, the highest among compared methods,
and competitive results on VSI-Bench, OST-Bench, and
MindCube-Tiny. These results highlight the importance
of combining online visual--geometric integration with
source-grounded historical evidence retrieval for spatial
reasoning over evolving RGB streams.

\section*{LLM Usage Statement}
\label{sec:llm-usage}

The research ideas and methodological design were developed
by the authors. ChatGPT (OpenAI) was used to assist with
literature exploration and manuscript editing, primarily
for language polishing and clarity refinement. ChatGPT's
image-generation and editing capabilities were also used
for iterative layout refinement and visual improvement
of selected figures. The LLM-based evaluator
\texttt{gpt-6.1-sol} was used to identify and assess
required information items in open-ended answers (OEA),
using a consistent evaluator prompt across models and
the item-level scoring protocol described in Appendix~B.

The authors designed the evaluation protocol, reviewed
the AI-assisted materials, and take full responsibility
for the research methodology, experimental implementation,
reported results, citations, and conclusions.

{
    \small
    \bibliographystyle{ieeenat_fullname}
    \bibliography{main}
}

\maketitlesupplementary

\appendix

\setcounter{figure}{0}
\setcounter{table}{0}
\setcounter{equation}{0}

\renewcommand{\thefigure}{S\arabic{figure}}
\renewcommand{\thetable}{S\arabic{table}}
\renewcommand{\theequation}{S\arabic{equation}}

\renewcommand{\theHfigure}{supp.\arabic{figure}}
\renewcommand{\theHtable}{supp.\arabic{table}}
\renewcommand{\theHequation}{supp.\arabic{equation}}

\section{Overview of Supplementary Material}
\label{app:overview}
Appendix~B provides implementation, training, and evaluation settings.
Appendix~C reports auxiliary cache-management and long-video diagnostics.
Appendix~D presents additional qualitative examples.

\section{Implementation and Evaluation Details}
\label{app:implementation}

\subsection{Architecture Details}
\label{app:architecture}

\noindent\textbf{Backbone and feature alignment.}
GeoRecall uses Qwen2.5-VL-3B-Instruct and causal StreamVGGT.
The visual encoder, visual merger, and geometry encoder are frozen;
the language model, geometric projector, CVGI, and QHER are trainable.
Geometry is aligned to the visual grid using a shared spatial-cropping
convention, discarding incomplete boundary patch groups. Adjacent
$2\times2$ geometry patches are concatenated along the channel dimension
and passed through a shared RMSNorm--Linear--GELU--Linear projector
into the Qwen hidden space. All block indices are zero-based.

\noindent\textbf{Historical evidence memory.}
The memory stores eight recent dense and eight historical compressed
observations, independently of both KV caches. When the recent bank
overflows, its oldest observation is transferred to the historical bank
using shared masked mean pooling over non-overlapping bins, each
containing at most $7\times7$ cells. Visual features and all three
geometry levels retain aligned coordinates, validity masks,
and source addresses.

When multiple historical candidates compete for eight slots, the policy
keeps the earliest and latest candidates and fills the remaining slots
by greedy farthest-point selection. The fixed distance metric equally
weights normalized visual cosine distance, $G^{17}$ geometric cosine
distance, and temporal separation normalized by the candidate span.
Descriptors are normalized spatial means computed before compression;
ties are broken in favor of the smaller source ID. Retention is
question-agnostic and cannot recover previously evicted observations.

\begin{table}[t]
\centering
\caption{Main architecture and memory configuration of GeoRecall.}
\label{tab:implementation_config}
\footnotesize
\setlength{\tabcolsep}{3.5pt}
\renewcommand{\arraystretch}{1.08}
\begin{tabularx}{\columnwidth}{@{}lX@{}}
\toprule
Configuration & Setting \\
\midrule
Language model & 36 blocks; hidden dimension 2,048 \\
Geometric feature layers & $G^{11}$, $G^{17}$, and $G^{23}$ \\
Spatial alignment & Align the geometry grid with the visual grid using a shared spatial-cropping convention; discard incomplete boundary regions without interpolation \\
Patch grouping & Concatenate each adjacent $2\times2$ geometry patch group along the channel dimension ($C\rightarrow4C$) \\
Shared projector & RMSNorm $\rightarrow$ Linear $\rightarrow$ GELU $\rightarrow$ Linear; maps features to the Qwen hidden dimension \\
Online injection & Apply a gated $G^{23}$ residual to the input visual embeddings; after Qwen blocks 0, 1, and 2, respectively, update the aligned visual tokens using $G^{11}$, $G^{17}$, and $G^{23}$ \\
Answer-time Readers & After Qwen blocks 1, 3, and 5, respectively, read $G^{11}$, $G^{17}$, and $G^{23}$ from the source observations associated with the Evidence Tokens, and update the Evidence Token representations \\
Reader width / attention heads & 512 / 8 \\
Retrieval-control dimension & 256 \\
Evidence Memory & Up to 8 recent dense and 8 historical compressed observations; historical grid up to $7\times7$ \\
Evidence tokens & Up to 8 per observation and 128 in total \\
\bottomrule
\end{tabularx}
\end{table}

\noindent\textbf{Historical reading.}
The controller concatenates separately normalized question and latest
visual-summary states and linearly projects them to 256 dimensions.
Scaled dot-product patch scores are normalized within each valid
spatial quadrant to form local Evidence Tokens; context tokens use
ordinary regional means. Empty regions are skipped. Frame scores combine
local and global relevance with relative observation-position features.
The resulting softmax priorities adjust Reader write gates without
removing retained frames or changing their source bindings.

Temporal features encode source--question, answer--source, and
answer--question gaps in observation-index units rather than seconds,
using signed-log values and their sine transforms. Readers attend only
to each token's source observation. Local attention assigns 90\% of
the weight to the regional distribution and 10\% to a uniform
distribution over all valid locations in that source observation;
context attention uses no regional bias. These soft regional preferences
preserve access beyond the token's own quadrant while enforcing
the source constraint.

\subsection{Training Datasets}
\label{app:training_data}

\noindent\textbf{Training data.}
The training mixture contains 1,592,833 samples: 1,002,166 from
Stream3D and 590,667 from VSI-590K. Stream3D contributes one CA-1M,
five ScanNet, and four ScanNet++ subsets covering camera motion,
environment measurement, object attributes, spatial relations, and
observation ordering. VSI-590K covers 12 question types organized
around size, direction, count, distance, and appearance order.
All architecture variants use the same data mixture.

\subsection{Architecture Comparison and Component Ablations}
\label{app:exp_ablation}

\noindent\textbf{Group A: Historical architecture comparison.}
Group A compares complete architectures A1--A4 on the full
Stream3D-Bench and VSI-Bench benchmarks, with all variants
trained for 49,801 steps. A1 is the full GeoRecall architecture,
combining online CVGI with source-constrained historical reading
through QHER. These comparisons evaluate alternative architectural
designs rather than controlled single-component interventions.

\noindent\textbf{A2: Multi-Level Routing.}
A2 uses a learnable See-Query to predict geometry-layer mixing
weights and residual strengths. Attention-based Readers then
inject the resulting mixed geometric features into visual
representations. In contrast, GeoRecall establishes explicit
layer-wise correspondence between geometric levels and Qwen
blocks, using visually conditioned gated residual injection.
Thus, A2 dynamically mixes geometric levels, whereas GeoRecall
preserves their correspondence to different representation depths.

\noindent\textbf{A3: Question-Conditioned Routing.}
A3 uses question representations to modulate online geometric
routing. In GeoRecall, online CVGI remains question-agnostic,
while explicit question conditioning is applied by QHER at
answer time to guide historical evidence reading. The distinction
is whether question information influences geometric fusion
during observation or conditions historical geometry access
during answer generation.

\noindent\textbf{A4: Historical Reading.}
A4 retrieves mixed geometry from a historical window through
Readers after Qwen blocks 4, 15, and 24. GeoRecall instead
combines online CVGI after blocks 0, 1, and 2 with
source-constrained QHER Readers after blocks 1, 3, and 5.
Each Evidence Token accesses geometry from its associated
source observation. A4 therefore differs in both the organization
of online and historical processing and the source constraints
imposed on historical geometry reading.

\noindent\textbf{Group B: Controlled component ablations.}
Group B evaluates component interventions on a fixed subset
of 500 questions, comprising 300 Stream3D-Bench and 200
VSI-Bench questions. B1 represents the full GeoRecall
architecture, while B2--B5 independently train the
corresponding ablated architectures. All five variants are
independently initialized from the same foundation-model
weights and trained for 8,000 steps using the same data
mixture, sampling order, effective batch size, optimizer
settings, learning-rate schedule, and random seed. No
variant reuses intermediate states from another run.

The fixed evaluation subset contains 100 questions from each
of the three Stream3D temporal settings and 25 questions from
each VSI task. Stratification accounts for answer type and
observation length, with sample selection determined by fixed
hashing without model predictions. Stream3D questions use
the common observation prefix ending at the annotated reference
Stop, where answering is forced. VSI questions use at most
32 uniformly sampled frames. All variants use greedy decoding,
a 512-token limit, random seed 17, and no BDKM.
The All score is the unweighted arithmetic mean of the two
benchmark scores. Group A and Group B results use different
evaluation scopes and are therefore not directly comparable.

\noindent\textbf{B2: Without online fusion.}
B2 bypasses the input-level $G^{23}$ residual and the subsequent
online geometric updates after Qwen blocks 0, 1, and 2.
Geometric encoding, historical Evidence Memory, Evidence
construction, and answer-time Geometry Readers remain enabled.
The current-question control pathway is also retained.
This intervention removes online geometric interaction while
preserving historical evidence access.

\noindent\textbf{B3: Without historical reading.}
B3 disables the insertion of historical Evidence Tokens and
their corresponding Geometry Reader updates, while retaining
online CVGI and the ordinary Qwen context. This intervention
removes the answer-time historical re-reading pathway without
removing online geometric fusion.

\noindent\textbf{B4: Single-level online geometry.}
B4 retains the original online injection positions, shared
projector, and gated residual interfaces, but replaces the
multilevel geometric inputs at all three online injection
positions with $G^{23}$. The input-level $G^{23}$ residual
is also retained. Historical memory and QHER continue to
use the original multilevel geometric sources. This intervention
isolates the contribution of geometric-level diversity in
online fusion.

\noindent\textbf{B5: Without explicit question conditioning.}
B5 replaces the explicit question state supplied to the reading
controller with a shared trainable vector $q_0$, yielding
\[
c_0=\phi_q(q_0,\operatorname{Pool}(V_{\mathrm{latest}})).
\]
The shared vector is optimized during training rather than
fixed to an arbitrary constant. The controller's operations
remain active, including patch scoring, frame prioritization,
Reader queries, spatial priors, and Reader write gates.
The original question is still provided to Qwen. Thus, B5
removes explicit semantic conditioning in the reading controller,
not all question information from the model. The Reader write
gates regulate how retrieved geometry updates Evidence Token
representations; they do not control historical-memory retention
or eviction.

\subsection{Evaluation Protocol}
\label{app:evaluation}

\noindent\textbf{Benchmark coverage and inputs.}
We evaluate on the full test sets: 10,037 Stream3D-Bench
questions, 5,130 VSI-Bench questions, 1,386 OST-Bench scenes
with 9,528 scored rounds, and 1,050 MindCube-Tiny questions.
All main evaluations use the same GeoRecall model with BDKM
disabled. Stream3D follows the official 1-FPS sequence with
Continue/Stop decisions. VSI uniformly samples at most 32
frames, including endpoints, and inserts the question before
the last frame. Both use $504\times392$ inputs and greedy
decoding with a 512-token answer limit. OST follows the
official per-round observations, retains previous answers,
and uses a 4,096-token limit per round. OST and MindCube
preserve the full field of view using aspect-ratio resizing
and centered padding to $504\times504$. MindCube uses all
2--4 supplied views and cyclically permutes answer options,
averaging normalized option log-probabilities mapped back
to the original choices across permutations. The OST adapter
uses causal-prefix replay for protocol compatibility; its
scores do not assess persistent-cache streaming efficiency.

\noindent\textbf{Scoring.}
NA and MCA follow the public benchmark scripts. Stream3D-VLM
and GeoRecall share an item-level OEA protocol using
\texttt{gpt-6.1-sol} with medium reasoning effort. Required
information items are fixed from the question and reference
answer before examining the prediction. Each question receives
the fraction of correctly answered items. Equivalent wording
and unit conversions are accepted. Scalar measurements use
a 15\% relative-error tolerance, rotation magnitudes an
absolute $10^\circ$ tolerance, and counts require exact
matches; object identity, direction, rotation sense, and
temporal order are checked separately. OEA is computed as
the unweighted mean of per-question item-level scores within
each temporal setting. The overall Stream3D-Bench Avg.
equally averages the nine NA, MCA, and OEA scores across
the three temporal settings. Other Stream3D reference results
retain their published OEA protocols.

OST follows its public hierarchical aggregation over eight
subgroups; Overall is not the arithmetic mean of the displayed
State, Info, and AO columns. MindCube computes Overall as
the question-count-weighted accuracy across Rotation
(200 questions), Among (600 questions), and Around
(250 questions). Additional baseline results combine public
reports and local evaluations; results obtained under
different OEA protocols are not strictly protocol-matched.

\noindent\textbf{Resource measurements.}
For GeoRecall, the TTFT Latency and End2End
Latency reported in the main results table correspond
to answer-stage TTFT and answer-generation latency,
respectively. Answer-stage TTFT is measured from the
committed Stop token to the first answer token.
Answer-generation latency uses the same starting point
and ends at the final answer token. Both include Evidence
construction and Geometry Readers but exclude preceding
observation processing. Thus, the reported End2End
Latency does not represent full video-level end-to-end
latency. CUDA peak allocated-memory statistics are reset
before each question and include resident model weights
and inference states. Table~\ref{tab:resource_records}
reports mean per-question resource measurements for
GeoRecall alongside published baseline records, which
retain their original measurement conventions. These
records are not paired hardware measurements.
Appendix~C reports current allocated memory and
per-observation latency during stress tests, excluding
initial CPU preprocessing and answer generation.

\begin{table}[t]
\centering
\caption{Resource statistics of different methods on Stream3D-Bench.}
\label{tab:resource_records}
\scriptsize
\setlength{\tabcolsep}{2pt}
\resizebox{\columnwidth}{!}{
\begin{tabular}{@{}lrrrr@{}}
\toprule
Method & TTFT (ms) & Completion (s) & Memory & Resolution\\
\midrule
GPT-4o & -- & -- & -- & 1296$\times$968 \\
GPT-5 & -- & -- & -- & 1296$\times$968 \\
LLaVA-Video-7B & 650 & 2.84 & 25.8 G & 1152$\times$768 \\
InternVL3-8B & 1034 & 5.65 & 40.5 G & 1024$\times$765 \\
InternVL3.5-8B & 1771 & 6.72 & 43.1 G & 1024$\times$765 \\
Qwen2.5-VL-7B & 325 & 3.16 & 40.0 G & 1008$\times$784 \\
Qwen2.5-VL-32B & 560 & 5.87 & 91.6 G & 1008$\times$784 \\
Qwen3-VL-8B & 150 & 1.56 & 28.8 G & 1024$\times$768 \\
Qwen3-VL-32B & 282 & 3.22 & 65.5 G & 1024$\times$768 \\
VideoLLM-online-8B (FT) & 120 & 1.29 & 28.4 G & 384$\times$384 \\
Qwen2.5-VL-7B (FT) & 106 & 0.53 & 36.8 G & 504$\times$392 \\
Stream3D-VLM-4B & 43 & 0.24 & 20.7 G & 504$\times$392 \\
Stream3D-VLM-8B & 62 & 0.39 & 36.6 G & 504$\times$392 \\
GeoRecall-4B (Ours) & 124.0 & 0.35 & 15.0 G & 504$\times$392 \\
\bottomrule
\end{tabular}
}
\end{table}

\section{BDKM Analysis and Frame-wise Evaluation}
\label{app:bdkm}

BDKM is a training-free cache-management extension applied during inference. It independently manages the Geometry KV and Qwen KV caches under separate capacity budgets, without merging their cache tensors or replacing the independent Historical Evidence Memory used by QHER. Since the two KV caches differ in representation and granularity, BDKM adopts different retention strategies for each. This section describes the cache-management mechanisms, the inference configuration and its resource and long-video diagnostics.

\subsection{Cache Management and Anchor Selection}
\label{app:bdkm_implementation}

\noindent\textbf{Geometry KV management.}
StreamVGGT maintains a causal Geometry KV cache as video observations arrive. BDKM applies attention-importance-based token pruning with a configured budget of 57,344 tokens, selectively retaining important historical geometric tokens rather than evicting entire observations. In the selected configuration, no additional anchor protection is applied to Geometry KV. Therefore, protecting a representative frame in Qwen KV does not guarantee that all geometric tokens from the same frame remain in Geometry KV.

\noindent\textbf{Qwen KV management.}
BDKM manages Qwen KV at the visual-frame-group level, evicting eligible historical frame groups while updating the associated cache positions and metadata. The configured total budget is 32,768 tokens, with a text quota of 4,096 tokens. Frame-group eviction preserves the grouping of visual tokens from each retained observation. In the selected configuration, representative frames selected using joint visual--geometric descriptors receive additional protection only in Qwen KV, changing retention priorities within the fixed budget rather than increasing cache capacity. The two KV caches are managed independently of the Historical Evidence Memory used by QHER. Pruning geometric KV tokens or evicting Qwen visual-frame groups does not, by itself, delete corresponding entries from this independently maintained evidence memory.

\noindent\textbf{Joint visual-geometric anchor selection.}
To select representative frames for additional Qwen KV protection, we combine visual and geometric descriptors. Visual descriptors are obtained by spatially pooling and normalizing Qwen image embeddings, while geometric descriptors are constructed from the corresponding pooled geometry features. Let $d_f^V$ and $d_f^G$ denote the normalized visual and geometric descriptors of frame $f$, respectively. Their joint representation is defined as
\begin{equation}
d_f^{GV}
=
\left[
\sqrt{1-\lambda}\,d_f^G;
\sqrt{\lambda}\,d_f^V
\right],
\qquad \lambda=0.5.
\label{eq:joint_anchor}
\end{equation}

Based on the joint descriptors, greedy farthest-point selection is performed using cosine distance over causally available candidates. The selected policy protects the first frame and retains up to eight representative anchors and four recent frames. Additional anchor protection is applied only to Qwen KV, while Geometry KV continues to follow its independent importance-based pruning policy.

\noindent\textbf{Cache-policy diagnostic.}
The 13-policy diagnostic uses a fixed Stream3D-VLM reference model, distinct from the GeoRecall evaluations below. It contains 80 VSI questions from 80 videos, with ten questions per task selected by fixed hashing without predictions. Each video supplies 256 uniformly sampled frames. Policies share weights, inputs, observation order, seed 17, and a 512-token answer limit, changing only the cache settings in Table~\ref{tab:bdkm_config_results}. QA scores range from 0 to 100. TTFT and E2E are median latencies, with TTFT measured from after Stop. E2E includes input loading, preprocessing, observations, question processing, and answering, excluding model loading and external scoring. Peak memory is the maximum single-question CUDA peak allocated memory. For Configurations 5--13, the first term specifies
the descriptor type used for anchor selection, while
the second identifies the KV cache receiving additional
anchor protection.

\begin{table}[t]
\centering
\caption{Cache and anchor configuration results on the VSI-Bench diagnostic subset.}
\label{tab:bdkm_config_results}
\scriptsize
\setlength{\tabcolsep}{2pt}
\renewcommand{\arraystretch}{1.08}
\resizebox{\columnwidth}{!}{
\begin{tabular}{@{}c l r r r r@{}}
\toprule
ID &
Cache Configuration / Anchor Protection &
VSI QA $\uparrow$ &
TTFT (ms) $\downarrow$ &
E2E (s) $\downarrow$ &
Peak Mem. (GiB) $\downarrow$ \\
\midrule
1 & Full Cache
& 57.22 & 40.87 & 68.18 & 50.71 \\

2 & Geometry KV Compression
& \textbf{62.22} & 40.76 & 52.35 & 20.62 \\

3 & Qwen KV Compression
& 54.86 & \textbf{30.68} & 67.93 & 49.59 \\

4 & Independent Dual-KV Compression
& 56.81 & 31.80 & \textbf{52.00} & 19.51 \\

5 & Geometry Anchors / Geometry KV
& 58.89 & 34.10 & 57.53 & 19.49 \\

6 & Geometry Anchors / Qwen KV
& 55.97 & 31.94 & \underline{52.10} & 19.51 \\

7 & Geometry Anchors / Both KV Caches
& 55.97 & 33.65 & 57.37 & 19.49 \\

8 & Visual Anchors / Geometry KV
& 58.75 & 33.84 & 57.38 & 19.49 \\

9 & Visual Anchors / Qwen KV
& 58.89 & \underline{31.54} & 52.26 & 19.51 \\

10 & Visual Anchors / Both KV Caches
& 57.22 & 33.74 & 57.61 & 19.49 \\

11 & Joint Visual-Geometric Anchors / Geometry KV
& 58.19 & 33.82 & 58.10 & 19.49 \\

\textbf{12} & \textbf{Joint Visual-Geometric Anchors / Qwen KV (Selected)}
& \underline{60.86} & 31.87 & 52.81 & 19.51 \\

13 & Joint Visual-Geometric Anchors / Both KV Caches
& 57.92 & 33.78 & 57.90 & 19.49 \\
\bottomrule
\end{tabular}
}
\end{table}

Among policies bounding both caches, joint anchors protected only in Qwen KV obtain the highest diagnostic QA score (60.86), compared with 56.81 for independent dual compression at the same 19.51 GiB peak-memory level. Geometry-only compression scores 62.22 but leaves Qwen KV unbounded. This diagnostic compares retention policies; the GeoRecall tests below evaluate the chosen bounded configuration.

\subsection{Resource Efficiency and Long-Video QA}
\label{app:bdkm_effectiveness}

\begin{figure}[t]
\centering
\includegraphics[width=\columnwidth]{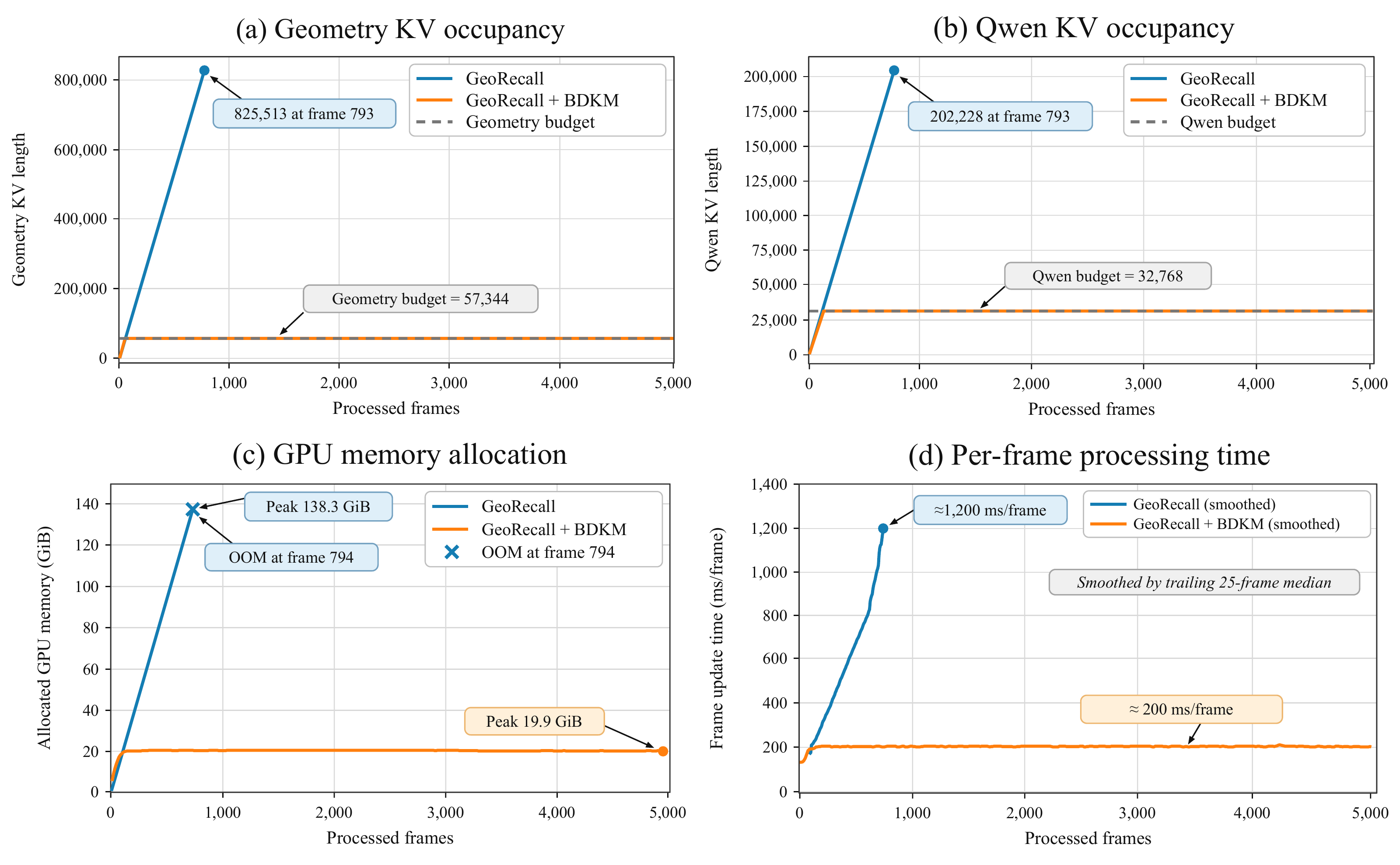}
\caption{Streaming stress test of GeoRecall with and
without BDKM over 5,000 frames. (a,b) Geometry
and Qwen KV occupancy; (c) GPU allocated memory;
(d) per-frame processing time. Without BDKM, GeoRecall
runs out of memory at frame 794, whereas GeoRecall
with BDKM completes the 5,000-frame test.
The processing-time curves use a trailing 25-frame
median.}
\label{fig:bdkm_resources}
\end{figure}

\noindent\textbf{Memory efficiency.}
We compare GeoRecall with Full Cache and BDKM on the same ordered sequence of 5,000 distinct observations uniformly sampled from a single video. This is a streaming stress test, not a sequence of 5,000 consecutive native frames. Both runs use identical model weights and inputs, differing only in cache policy. As shown in Figure~\ref{fig:bdkm_resources}, BDKM bounds Geometry KV and Qwen KV within their respective budgets of 57,344 and 32,768 tokens, whereas Full Cache exhibits continuous cache growth as observations accumulate.

Under Full Cache, Geometry KV and Qwen KV reach 825,513 and 202,228 tokens, respectively, at observation 793, followed by an out-of-memory (OOM) failure at observation 794. BDKM completes all 5,000 observations with a peak allocated memory of 19.9 GiB, compared with 138.3 GiB for Full Cache before its failure. These are run-level peak measurements over different completed sequence lengths, rather than a matched same-prefix memory comparison. Processing latency also differs: Full Cache increases to approximately 1,200 ms per observation near termination, whereas BDKM maintains approximately 198--200 ms per observation. Overall, these results show that BDKM bounds cache growth and enables substantially longer stream processing under limited GPU memory.

\noindent\textbf{Long-video spatial QA evaluation.}
We compare 32-frame sampling and native frame-wise processing on a randomly sampled diagnostic subset of 80 questions from 30 videos. The three video-length groups contain 10 videos each and 27, 27, and 26 questions, respectively. Both settings use the same GeoRecall weights and selected BDKM configuration.

The 32-frame inputs remain within both cache budgets and incur no frame eviction. This setting inserts the question before the last sampled frame and uses four-frame prefill blocks, whereas native frame-wise processing inserts the question after the video and uses one-frame blocks. Thus, the comparison changes not only observation density but also question placement and processing granularity. QA scores are mean per-question scores ranging from 0 to 100, including partial credit. The overall scores are question-count-weighted means computed from unrounded per-question scores. Table~\ref{tab:long_video_qa} reports the diagnostic results, with $\Delta$ QA defined as frame-wise QA minus 32-frame QA. This subset is intended for diagnostic analysis and does not represent full-benchmark performance.

\begin{table}[t]
\centering
\caption{Spatial QA comparison between 32-frame
sampling and native frame-wise processing on a
randomly sampled long-video diagnostic set.
$\Delta$QA is computed as Frame-wise QA minus
32-frame QA.}
\label{tab:long_video_qa}
\footnotesize
\setlength{\tabcolsep}{2.8pt}
\renewcommand{\arraystretch}{1.15}
\resizebox{\columnwidth}{!}{
\begin{tabular}{@{}lcccc@{}}
\toprule
Video length (frames) &
Videos / Questions &
32-frame QA &
Frame-wise QA &
$\Delta$ QA \\
\midrule
1,000--3,499
& 10 / 27 & 58.89 & 45.19 & $-13.70$ \\

3,500--6,499
& 10 / 27 & 42.96 & 52.22 & $+9.26$ \\

6,500--12,000
& 10 / 26 & 41.15 & 16.15 & $-25.00$ \\
\midrule
\textbf{Overall}
& \textbf{30 / 80}
& \textbf{47.75}
& \textbf{38.13}
& $\mathbf{-9.62}$ \\
\bottomrule
\end{tabular}
}
\end{table}

\noindent\textbf{Analysis.}
Frame-wise processing improves QA by 9.26 points in the intermediate-length group, but decreases by 13.70 and 25.00 points in the shorter and longest groups, respectively. Overall, frame-wise processing scores 38.13, compared with 47.75 for 32-frame sampling. Although the questions are randomly sampled, the two execution settings differ in question placement and prefill granularity, so this comparison does not isolate the effects of frame density or processing strategy. Moreover, both settings use BDKM, so the QA results do not establish its contribution to answer accuracy. Together, the diagnostics show that BDKM enables bounded long-stream processing, but does not guarantee improved spatial QA performance.

\section{Additional Qualitative Results}
\label{app:qualitative}

These nine cases complement the aggregate results in the main
paper by illustrating object identification, inter-object
spatial relations, and object-to-camera direction and distance
estimation. Each case compares the reference answer with
predictions from Stream3D-VLM Release and GeoRecall, together
with the source observations used by QHER. Only sources with
at least 10\% priority are displayed, without renormalization.
Priority reflects evidence-reading weights rather than
localization confidence or causal contribution.


\qualcasefigure
{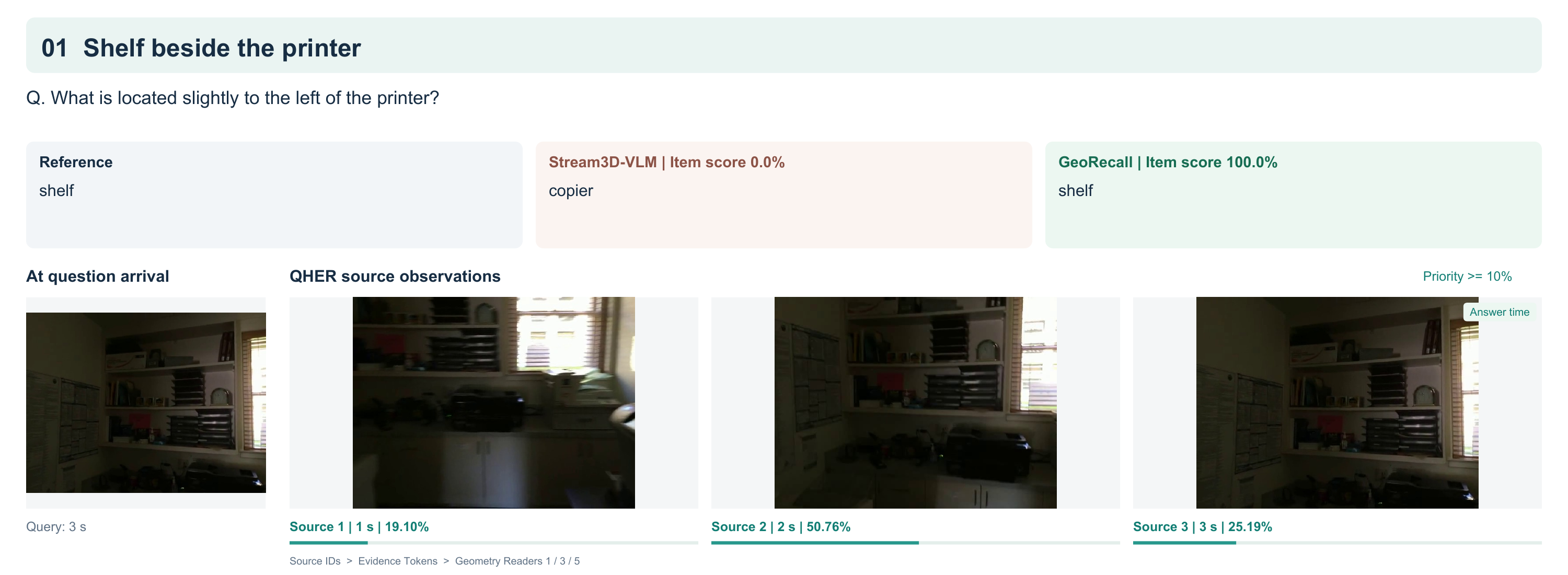}
{Case 1: Shelf beside the printer. The highest-priority source (50.76\%) shows the printer area. GeoRecall identifies the shelf, whereas Stream3D-VLM Release predicts a copier.}
{fig:qual_shelf_printer}


\qualcasefigure
{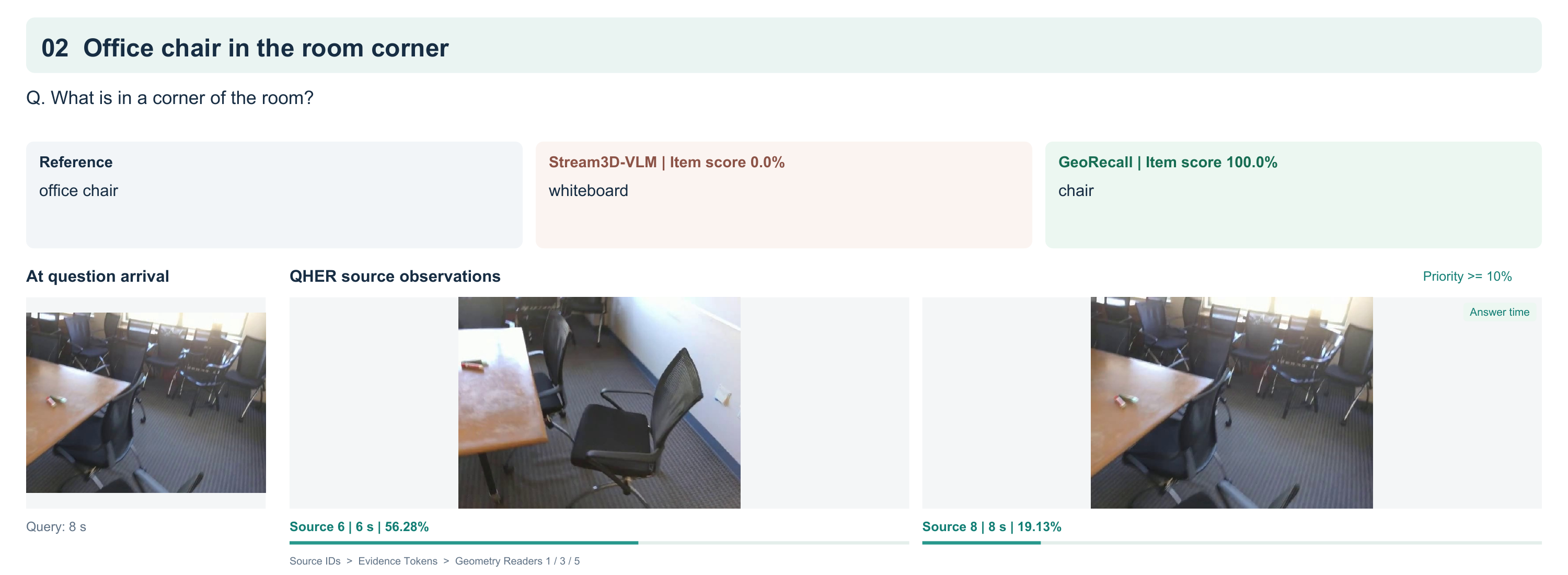}
{Case 2: Office chair in the room corner. Retained views show chairs beside the table and wall. GeoRecall identifies the office chair, whereas Stream3D-VLM Release predicts a whiteboard.}
{fig:qual_office_chair_corner}


\qualcasefigure
{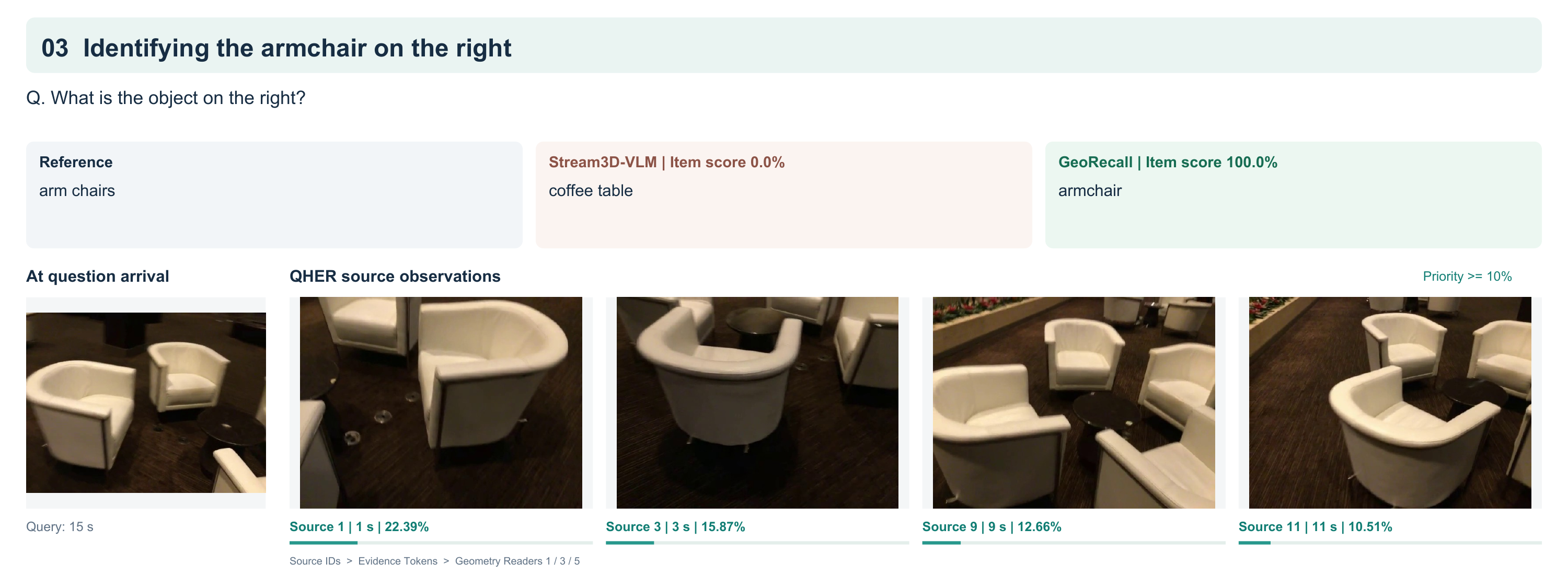}
{Case 3: Identifying the armchair on the right. Source views show its curved back and armrests. GeoRecall identifies the armchair, whereas Stream3D-VLM Release predicts a coffee table.}
{fig:qual_armchair}


\qualcasefigure
{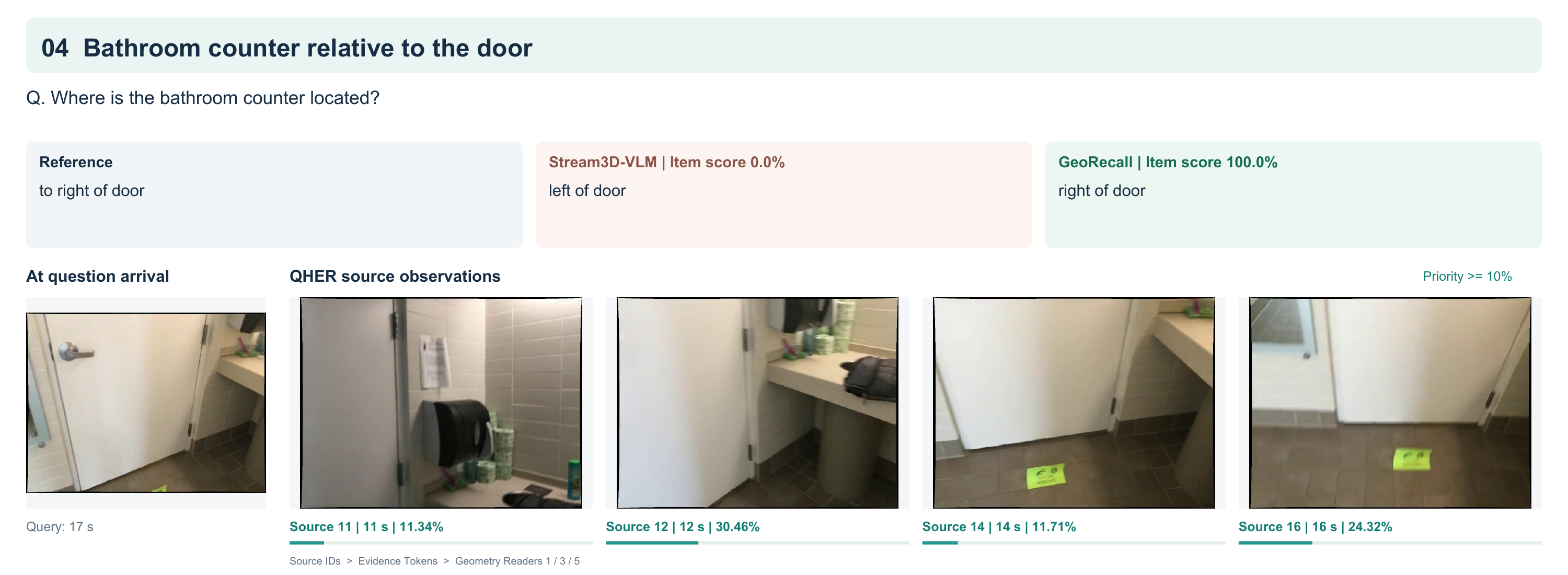}
{Case 4: Bathroom counter relative to the door. Retained views show both objects. GeoRecall places the counter to the right of the door, whereas Stream3D-VLM Release predicts the opposite direction.}
{fig:qual_counter_door}


\qualcasefigure
{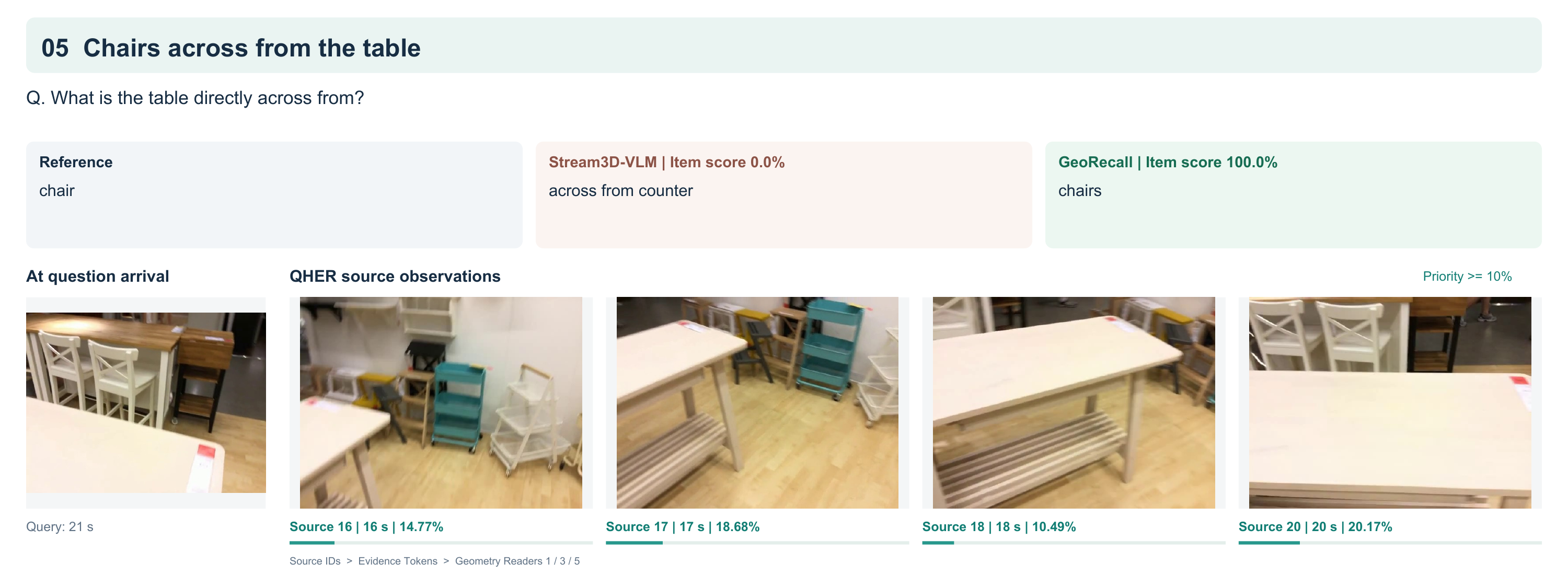}
{Case 5: Chairs across from the table. Later source views show the spatial arrangement. GeoRecall identifies the chairs, whereas Stream3D-VLM Release gives an incorrect spatial description.}
{fig:qual_chairs_table}


\qualcasefigure
{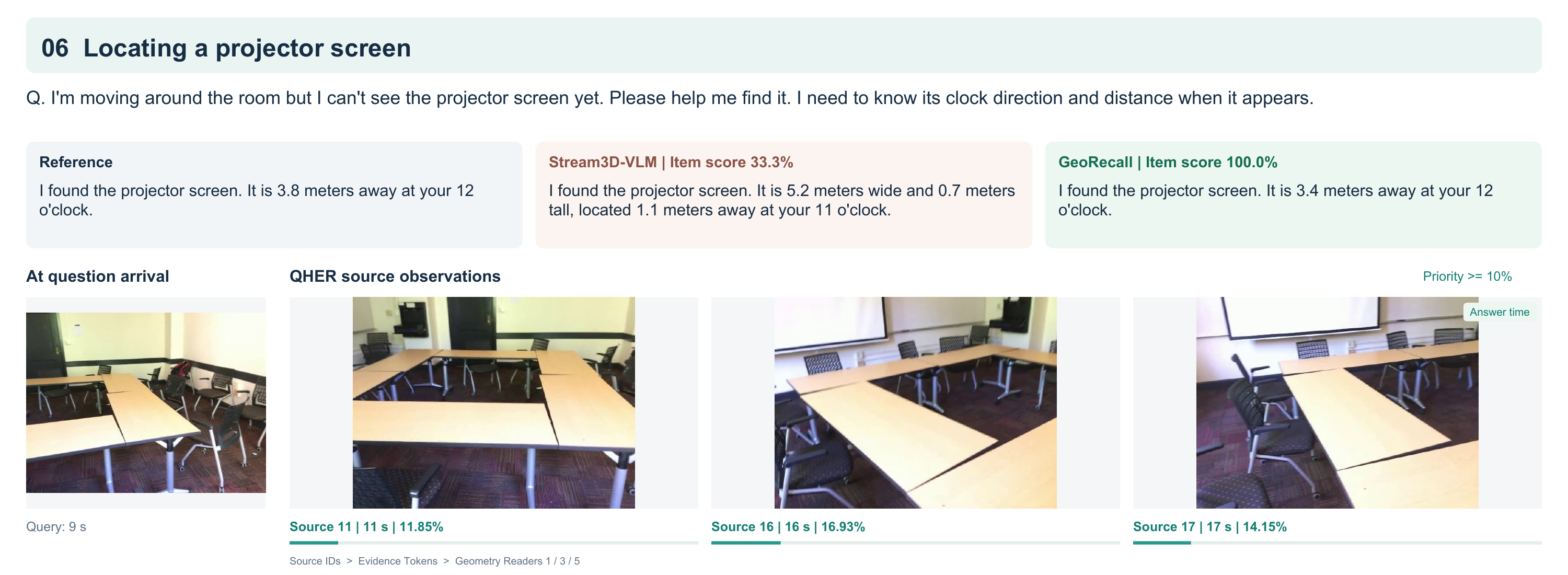}
{Case 6: Locating a projector screen. After question arrival, GeoRecall predicts 12 o'clock and 3.4 m, within the 15\% relative-error tolerance of the 3.8 m reference distance. Stream3D-VLM Release predicts 11 o'clock and 1.1 m.}
{fig:qual_projector_screen}


\qualcasefigure
{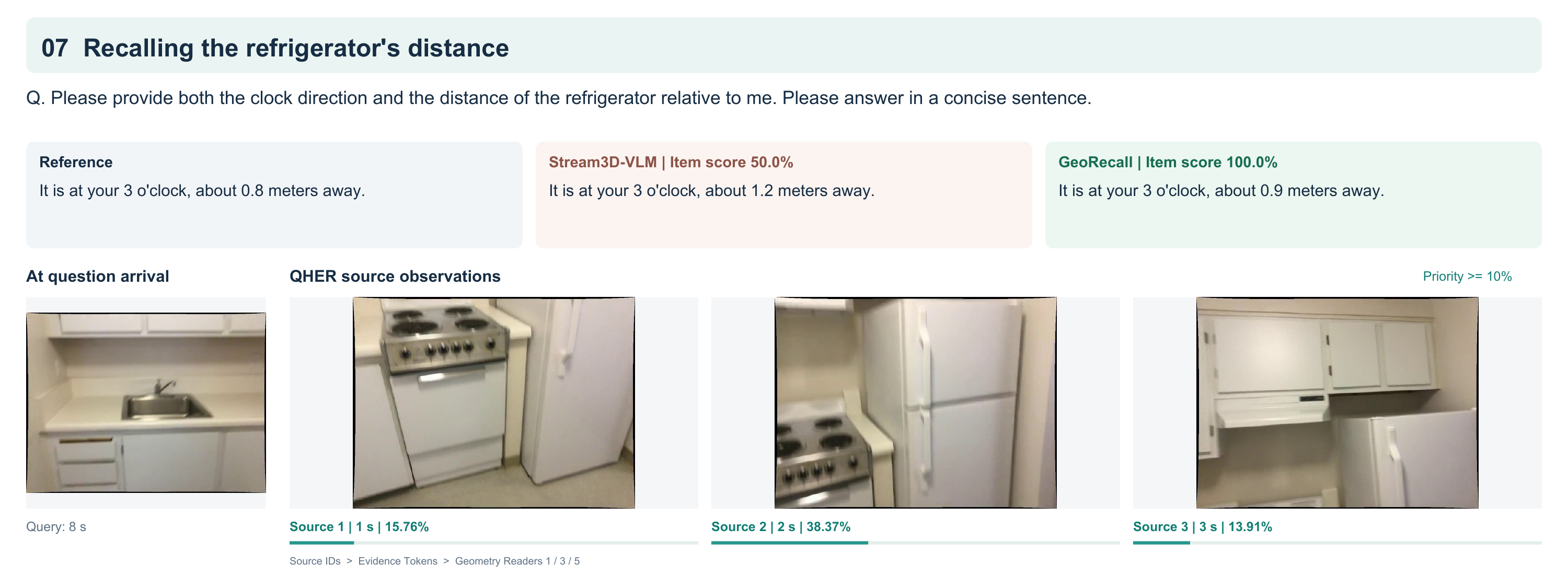}
{Case 7: Recalling the refrigerator's distance. Earlier views retain the refrigerator after the camera turns toward the sink. Both predict 3 o'clock; GeoRecall's 0.9 m falls within the 15\% tolerance of the 0.8 m reference, whereas Stream3D-VLM Release's 1.2 m does not.}
{fig:qual_refrigerator}


\qualcasefigure
{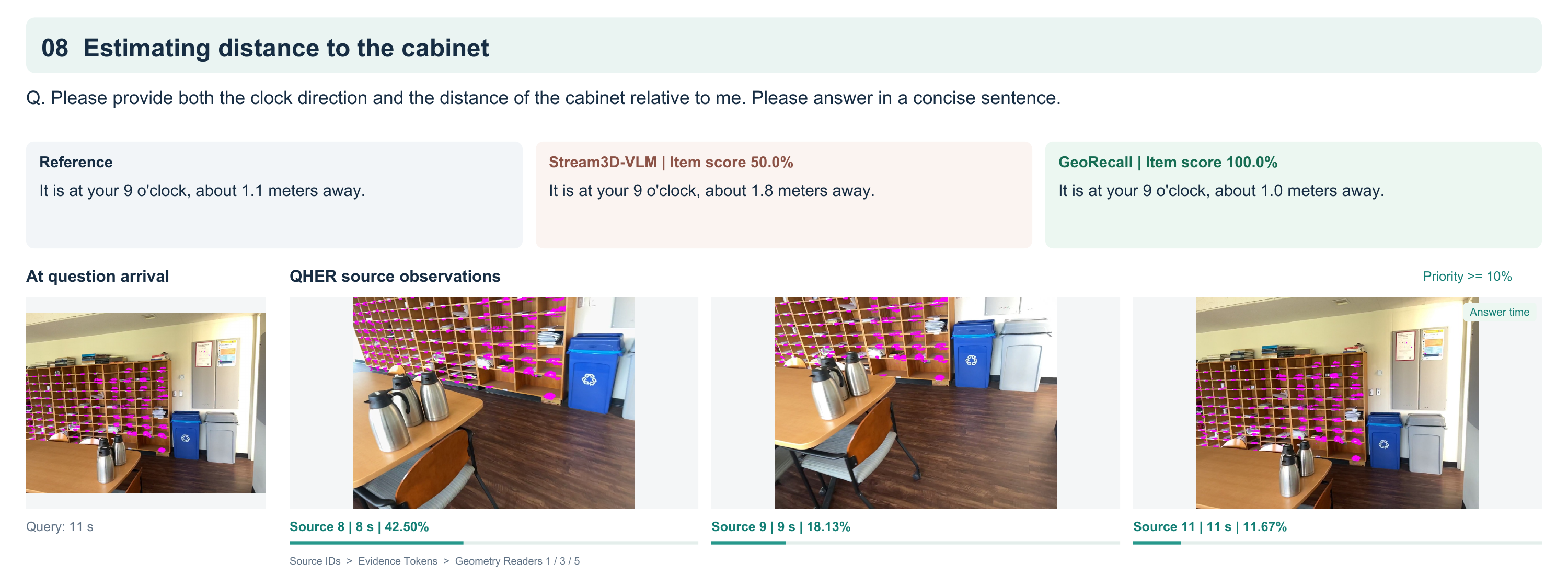}
{Case 8: Estimating distance to the cabinet. The highest-priority source receives 42.50\% priority. Both predict 9 o'clock; GeoRecall's 1.0 m falls within the 15\% tolerance of the 1.1 m reference, whereas Stream3D-VLM Release's 1.8 m does not.}
{fig:qual_cabinet}


\qualcasefigure
{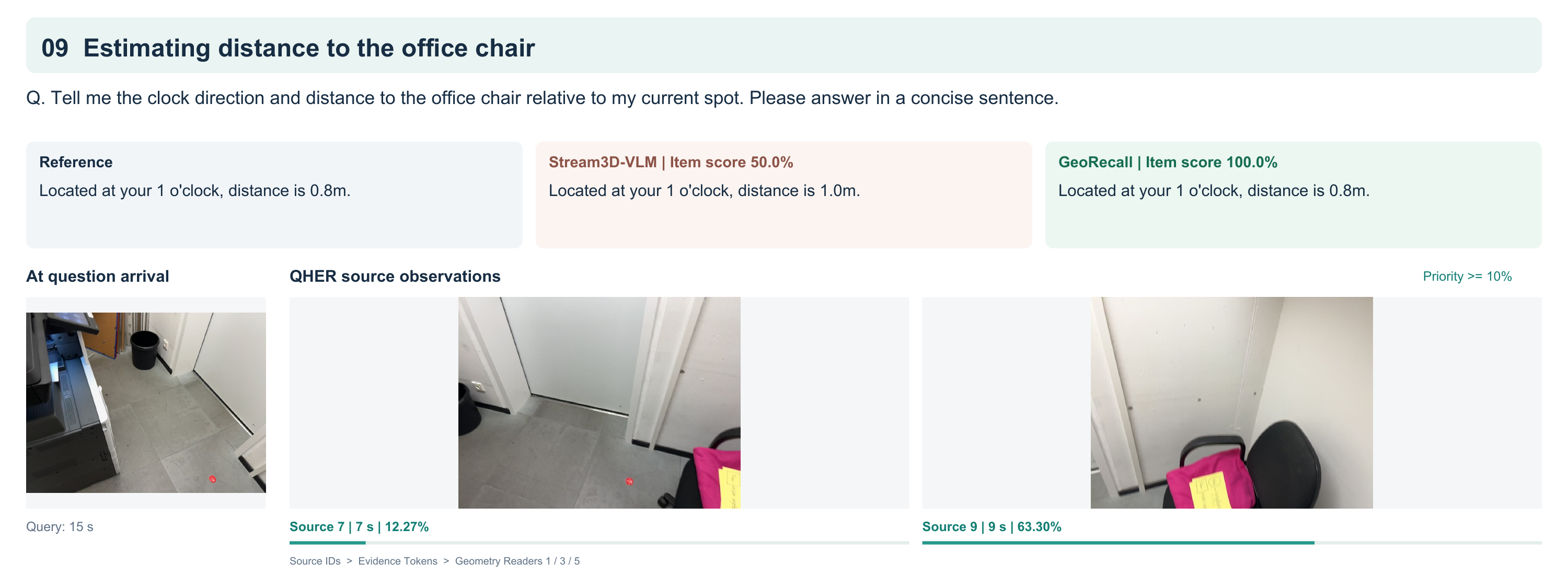}
{Case 9: Estimating distance to the office chair. The 9 s source receives 63.30\% priority. Both predict 1 o'clock; GeoRecall predicts the reference distance of 0.8 m, whereas Stream3D-VLM Release's 1.0 m exceeds the 15\% tolerance.}
{fig:qual_office_chair_distance}


\end{document}